\documentclass{article}

\usepackage[final]{neurips_data_2022}

\usepackage[utf8]{inputenc} 
\usepackage[T1]{fontenc}    
\usepackage{hyperref}       
\usepackage{url}            
\usepackage{booktabs}       
\usepackage{amsfonts}       
\usepackage{nicefrac}       
\usepackage{microtype}      
\usepackage{xcolor}         
\usepackage[flushleft]{threeparttable} 
\usepackage[pdftex]{graphicx} 
\usepackage{kotex}
\usepackage{makecell}
\usepackage{hyperref}
\usepackage{natbib}
\usepackage{amsmath}
\usepackage{xcolor}
\hypersetup{
    colorlinks,
    linkcolor={black!50!black},
    citecolor={black!50!black},
    urlcolor={black!80!black}
}

\usepackage[utf8]{inputenc}

\usepackage{multicol}

\usepackage{lineno}

\newcommand{\ours}{\textsc{LBox Open}}

\newcommand{\oursm}{\textsc{LCube}} 

\newcommand{\casename}{\textsc{Case name}}
\newcommand{\statute}{\textsc{Statute}}
\newcommand{\ljpcriminal}{\textsc{LJP-criminal}}
\newcommand{\ljpcivil}{\textsc{LJP-civil}}
\newcommand{\summarization}{\textsc{Summarization}}
\newcommand{\precedent}{\textsc{Precedent corpus}}
\newcommand{\testex}{test2}
\newcommand{\ditto}{\texttt{"}}

\newcommand{\casenamep}{\textsc{Case name+}}
\newcommand{\statutep}{\textsc{Statute+}}
\newcommand{\summarizationp}{\textsc{Summarization+}}

\title{A Multi-Task Benchmark for Korean Legal Language Understanding and Judgement Prediction}

\author{
  Wonseok Hwang$^a$ \quad
  Dongjun Lee$^a$ \quad 
  Kyoungyeon Cho$^a$ \quad
  Hanuhl Lee$^{a}$ \quad 
  Minjoon Seo$^{a, b}$ \quad 
  \\
  $^a$LBox \\
  $^b$KAIST \\
  \texttt{\{wonseok.hwang, dongjun.lee, kycho, leehanuhl\}@lbox.kr} \\
  \texttt{minjoon@kaist.ac.kr}\\
}

\begin{document}

\maketitle

\begin{abstract}
The recent advances of deep learning have dramatically changed how machine learning, especially in the domain of natural language processing, can be applied to legal domain. However, this shift to the data-driven approaches calls for larger and more diverse datasets, which are nevertheless still small in number, especially in non-English languages.
Here we present the first large-scale benchmark of Korean legal AI datasets, \ours, that consists of one legal corpus, two classification tasks, two legal judgement prediction (LJP) tasks, and one summarization task.
The legal corpus consists of 147k Korean precedents (259M tokens), of which 63k are sentenced in last 4 years and 96k are from the first and the second level courts in which factual issues are reviewed.
The two classification tasks are case names (11.3k) and statutes (2.8k) prediction from the factual description of individual cases.
The LJP tasks consist of (1) 10.5k criminal examples where the model is asked to predict fine amount, imprisonment with labor, and imprisonment without labor ranges for the given facts, and (2) 4.7k civil examples where the inputs are facts and claim for relief and outputs are the degrees of claim acceptance.
The summarization task consists of the Supreme Court precedents and the corresponding summaries (20k). We also release realistic variants of the datasets by extending the domain (1) to infrequent case categories in case name (31k examples) and statute (17.7k) classification tasks,  and (2) to long input sequences in the summarization task (51k).
Finally, we release \oursm, the first Korean legal language model trained on the legal corpus from this study. 
Given the uniqueness of the Law of South Korea and the diversity of the legal tasks covered in this work, we believe that \ours\ contributes to the multilinguality of global legal research.
\ours\ and \oursm\ will be publicly available\footnote{\url{https://github.com/lbox-kr/lbox-open}}.
\end{abstract}

\section{Introduction}
The application of artificial intelligence in a legal domain has a long history, dating back to 1980s \citep{ashley2017legal_analytics}. Although previous legal expert systems based on rules and domain knowledges showed some useful results on certain legal areas such as settlement decisions of product liability disputes \citep{waterman1981legaldecision}, such system could not be easily extended beyond narrow domains.
More recently, the advances in deep learning are fundamentally changing how and where machine learning can be applied in legal domain, such as 
legal judgement prediction \citep{chalkidis2019aclLJP,ma2021sigarLJPRealCourtSetting,zhong2020aaai_ljp_qa_interpretable,xiao2018cail2018},
legal content generation \citep{wu2020emnlp_court_view_gen, tonguz2021NLLUautomatingPatent}, 
legal text classification \citep{chalkidis2021multieurlex, chalkidis2022FACL_doc_class_data_imblance,hendrycks2021cuad}, 
legal event detection \citep{yao2022FACLlevenEventDetection}, 
legal information extraction \citep{hong2021nllu_ie_dialogue}, 
legal contract review \citep{hendrycks2021cuad, koreeda2021contractnli-dataset},
and question answering \citep{khazaeli2021nllu_qa,zhong2019jecqa,duan2019cjrc}.

This shift of the paradigm to data-driven approaches necessitates the use of large-scale datasets for training machine learning systems. In response to this, legal research groups worldwide have worked on various legal AI datasets across multiple languages recently \citep{paul2022lesicin,kapoor2022facl_hindi_legal_corpus,yao2022FACLlevenEventDetection,glaser2021nllu_german_court_summarization,niklaus2021nllu_swiss_ljp,chalkidis2022acl_lexglue,rossi2021verbcl,chalkidis2022facl_fairlex,louis2022acl_belgian_statutory_retrieval,rabelo2020coliee}.

However, despite the fact that Korea has a legal industry market size of $\sim$4.8B USD\footnote{the size is estimated based on the value-added tax base declaration amount \url{https://m.lawtimes.co.kr/Content/Article?serial=163743)}} it still lacks large-scale corpora, benchmark datasets, and language models in the domain of Korean Law. 
AI-hub, an organization run by Korean government, released some Korean legal AI datasets in 2020\footnote{{\url{https://aihub.or.kr/aidata/7974}}} but they consist of a relatively small corpus (obtained from 6k precedents) and a single named entity recognition task, which lack both the scale and the diversity of the tasks.

In this work, we release \ours, the first large-scale Korean legal AI benchmark that consists of six datasets: (1) a large-scale legal precedent corpus (\precedent), (2) two classification tasks (\casename, \statute), (3) two legal judgement prediction tasks (\ljpcriminal, \ljpcivil), and (4) one summarization task (\summarization).
\precedent\ consists of 147k precedents (259M tokens) of which 63k are sentenced in last 4 years and 96k are from the first and the second level courts.
\casename\ consists of 11.3k facts and case name pairs and \statute\ includes 2.8k facts and the corresponding statutes pairs. In both tasks, a model needs to use the facts to predict the corresponding case names or statutes.
We also release the extended datasets \casenamep\ (31k), and \statutep\ (17.7k) by including infrequent case categories.
\ljpcriminal\ consists of three subtasks: fine amount, imprisonment with labor or imprisonment without labor range predictions. In each task, a model is expected to use the facts to predict the judgement results. The dataset consists of 11k examples from 7 case categories.
In \ljpcivil, a model is asked to predict the claim acceptance degree for the given facts and the gist of the claim. The range is determined based on how much claimed money from plaintiffs is approved by the judge.
\summarization\ consists of 20k precedents from the Korean Supreme Court and the corresponding summaries. In addition, we also release \summarizationp\ (51k) by including examples with long sequences. 
The comparison of \ours\ to previous studies are presented in Appendix \ref{sec: related works}.

We also release the first large-scale Korean legal language model \oursm\ (124M, 345M) pre-trained using \ours.
Comparisons with other language models on our benchmark affirm that a domain-specific corpus is necessary to achieve competent performance on the tasks. For the text classification tasks, \oursm, which is a decoder-only model based on GPT-2, shows comparable performance with mt5~\citep{raffle2019t5}, a competitive encoder-decoder language model with larger size. On the other hand, on the summarization task, \oursm\ does not seem to have advantage over other models. 

Given the uniqueness of the Law of South Korea and the diversity of the legal tasks covered in this work, we believe that \ours\ not only contributes to the multilinguality of global legal research but also serves as a good example for creating a benchmark for legal language understanding and judgement prediction in other languages.
Both \ours\ and \oursm\ will be publicly available.

\section{Background}

\begin{figure}[t]
\centering
\includegraphics[width=0.5\textwidth]{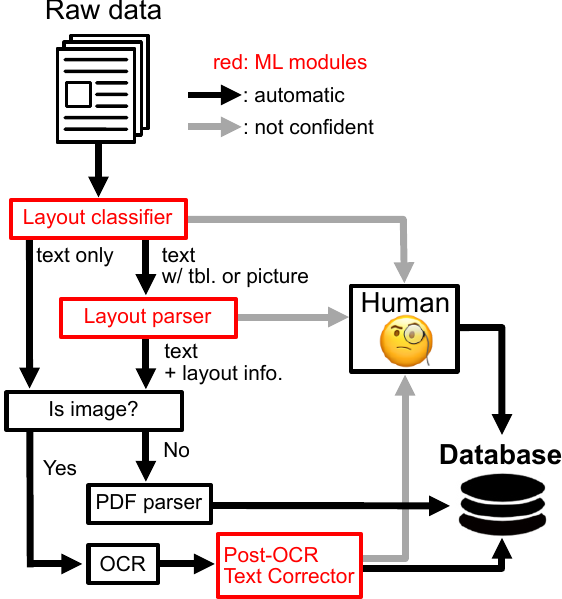}
\caption{Precedents structuring pipeline. The raw data is formatted as images or PDF.} 
\label{fig_pipeline}
\vspace{-4mm}

\end{figure}

\subsection{Korean legal system}

Korean court system is based on the three-tiered justice system, which is composed of district courts, the high courts, and the Supreme Court. The first two trials debate both the factual and legal issues of cases whereas the final trial only discusses the legal issues.

Korean legal system is rooted in civil law system. Unlike the common law system, which is widely accepted in the U.S. and the U.K., codified statute are of primary importance in civil law system and precedents only play supplementary roles. Therefore the doctrine of stare decisis, meaning that courts should adhere to precedent in making their decisions, is not institutionally adopted in Korea. Court Organization Act states that a higher court's decision on a matter only binds the lower court on the particular matter\footnote{Article 8, Korean Court Organization Act, \url{https://elaw.klri.re.kr/kor_service/lawView.do?hseq=55374&lang=ENG}}. In practice, however, higher courts' decisions, especially those of the Supreme Court, are typically followed by lower courts as statutes alone cannot cover whole complex social phenomena. 

\subsection{Korean precedent} \label{sec: korean precedent}
\paragraph{Structure of Korean precedent}
Korean precedents consist of five main parts, (1) meta information (such as case number, plaintiffs, defendants, and sentencing date) (2) gist of claim from plaintiffs in a civil case, (3) ruling, and (4) reasoning section where (4-1) facts, (4-2) claims, (4-3) reasoning, and (4-4)  decisions are loosely separated. The example of precedent is shown in Appendix \ref{sec: precedent example}.

\paragraph{The redaction process} 
The Korean government follows the official redaction process to avoid a possible risk of invading
privacy. By the redaction process, any data that may reveal personal information are anonymized
before being open to public. 
Data subjected to anonymization are described in Appendix \ref{sec: redaction_rule}.

\paragraph{Precedent disclosure status} According to Civil and Criminal Procedure Act, the courts are making their judgement decisions public via online service\footnote{https://www.scourt.go.kr/portal/information/finalruling/guide/index.html} provided by the Supreme Court of Korea. Also the Korean precedents are officially not protected by the copyright law\footnote{COPYRIGHT ACT, Article 7, \url{https://law.go.kr/LSW/lsInfoP.do?lsiSeq=192474&viewCls=engLsInfoR&urlMode=engLsInfoR\#0000},\texttt{  https://law.go.kr/법령/저작권법/제7조}}.
Despite this, the accessibility of the Korean precedents to the general public is still contentious in that the courts only provide final decisions that are irrevocable while charging a 1,000 won fee per a document. The more discussions on this topic is presented in Appendix \ref{sec: precedent disclosure status}.

\section{\ours\ datasets}

\subsection{Structuring raw data} \label{sec: structuring}
Korean precedents have been released by the Korean government in two formats: (1) document images, or (2) PDF. In each page, figures and tables are mixed with plain text, which makes parsing non-trivial. To automatically process them, we built our custom data engineering pipeline (Fig. \ref{fig_pipeline}). 
To process raw data, we first use \texttt{Layout classifier} based on ResNet \citep{he2016resnet} that classifies each page into \texttt{text only} or \texttt{text w/ tbl or pictures}. \texttt{Layout classifier} was trained using 300k training examples consisting of document image and corresponding label pairs.
Next, we use \texttt{Layout parser} that is based on Mask-R-CNN \citep{he2017maskrcnn,wu2019detectron2,Zhong2019publaynet} to segment any non-textual component from pages. 
Next, if the page is in PDF format, we extract text and use a custom rule-based parser before sending them to the database. When the page is in image format, we use a proprietary OCR engine\footnote{We use a combination of Google's (\url{https://cloud.google.com/vision}) and NAVER Clova's (\url{https://clova.ai/ocr}) OCR APIs.} to extract text segments and their coordinates. Then we use our custom language model to correct OCR errors before sending it to the database. The model uses the Korean character level transformer as backbone \footnote{\url{https://github.com/monologg/KoCharELECTRA}}. 
When the confidence scores from the ML modules are smaller than certain thresholds, corresponding pages are processed manually.
Final computer readable documents are saved in JSON format that includes (1) meta information such as case name, sentencing dates, and names of attendees, (2) ruling, (3) gist of claim, (4) appeal, (5) reasoning body that consists of facts, claims, reasoning, and decision of judges.
The further details about the data engineering pipeline are explained in Appendix \ref{sec: pipeline}.

\begingroup
\setlength{\tabcolsep}{1pt} 
\renewcommand{\arraystretch}{0.6} 

\begin{table}[t!]
\tiny
  \caption{Comparison of precedent corpus. \texttt{glaw} indicates the statistics of the set of precedents we retrieved from LAW OPEN DATA.}.
  \label{tbl_precedent_corpus}
  \centering
  \begin{threeparttable}
  \begin{tabular}{llll}
    \toprule
    Name     & \ours & glaw & ai-hub \\
    \midrule
    \midrule
    \# of precedents   & 146,791 & 82,258 & 5,958   \\ 
    \midrule
    \# of tokens$^\dagger$  & 259M & 157M & 24M       \\
    \midrule
    \# of trials from first and second level courts     
    &  \makecell[l]{56,557 (1st),\\39,570 (2nd)} 
    &\makecell[l]{11,278 (1st),\\24,085 (2nd)} 
    & 2,218 (1st, 2nd)  \\
    \midrule
    \# of precedents sentenced in last 4 years (2018--2021) & 63,439 & 4,416 & 692 \\
    \midrule
    \# of precedents of each case type 
    & \makecell[l]{58,932 (civil),\\54,624 (criminal),\\ 25,007 (administration),\\ 3,266 (patent),\\1,237 (family suit),\\3,725 (etc)} 
    & \makecell[l]{36,546 (civil),\\18,912 (criminal),\\20,824 (administration),\\3,219 (patent),\\1,210 (family suit),\\1,547 (etc)} 
    & - \\
    \midrule
    \# of major case categories$^*$ &  1,161 & 766 & 59 \\
    \bottomrule
  \end{tabular}
  \begin{tablenotes}[]
  \item $\dagger$ Based on KLUE \citep{park2021klue} tokenizer.
  \item $*$ The categories with at least 10 precedents in the corpus.
  \end{tablenotes}
  \end{threeparttable}
\end{table}
\endgroup

\subsection{Datasets} \label{sec: datasets}
\paragraph{\precedent}
The AI hub, an organization run by the Korean government, released a legal corpus in 2020. Although this has been a good stepping stone for legal AI research, it consists of only 6k precedents. As approximately 1.5 million Korean precedents are created every year\footnote{\url{https://www.scourt.go.kr/portal/justicesta/JusticestaListAction.work?gubun=10}}, it is difficult to expect the released precedents can cover various legal activities. 
To subside this limitation of the previous work, we release \precedent\ consisting of 147k precedents (259M tokens) that includes 96k precedents from the first and second level trials and 63k recent precedents. We retrieve 82k Korean precedents raw data from LAW OPEN DATA\footnote{\url{https://www.law.go.kr/LSO/main.do}} and additional 65k precedents from our internal database. The 57\% of precedents retrieved from LAW OPEN DATA consist of the trials of the Supreme Court where the factual issues of cases are not discussed. As statutes cannot comprehensively cover diverse aspects of complex social phenomena, precedents are used to cover the empty spaces of the statutes and thus the precedents from the first and the second trials discussing factual issues are most important to legal practitioners. In this regard we include additional 65k precedents from our internal database where 94\% of them are from first and second trials from recent 4 years (2018-2021). 
The difference between \precedent\ and other corpus is summarized in Table~\ref{tbl_precedent_corpus}.

\begingroup
\setlength{\tabcolsep}{0.5pt} 
\renewcommand{\arraystretch}{0.6} 

\begin{table}[t!]
\centering
\begin{threeparttable}

\scriptsize
  \caption{Task overview.}
  \label{tbl_tasks}
  \centering
  \begin{tabular}{llllcccl}
    \toprule
    Name     &  Task Type  & Input & Output & \# of case categories &\# of classes &  \makecell{Size \\train:valid:test:\testex$^*$}  \\
    \midrule
    \midrule
    \casename &  Classification & facts & case name & 100 &  100 &  8,000:1,000:1,000:1,294 \\
    \casenamep &  \ditto & \ditto & \ditto & 603 &  603 &  22,494:3,999:-:4,790 \\
    \midrule
    \statute   &  Classification & facts & statutes & 46 & 188 &  2,208:276:276:538  \\
    \statutep   &  \ditto & \ditto & \ditto & 434 & 1,015 &  13,317:2,276:-:2,137 \\
    \midrule 
    \ljpcriminal   & \makecell[l]{Legal judgement\\prediction} & facts & \makecell[l]{fine\\imprisonment w/ labor\\imprisonment w/o labor}& 7 & \makecell[c]{(5, 6, 6)$^{a}$\\or\\ (3, 3, 3)$^{b}$} & 8,400:1,050:1,050:928\\
    \midrule
    \ljpcivil   &  \makecell[l]{Legal judgement\\prediction} & \makecell[l]{facts\\gist of claim} & the claim acceptance degree &4 & 3 or 13$^b$ & 3,742:467:467:403 \\
    \midrule
    \summarization   &  Summarization & \makecell[l]{ruling\\reasoning section} & summary & &- & 16,000:2,000:2,000:- \\
    \summarizationp   &  \ditto & \ditto & \ditto & &- & 40,892:5,111:5,111:- \\

    \bottomrule
  \end{tabular}
  \begin{tablenotes}[]
  \item $*$ An additional test set made from precedents not included in \precedent. Built for the fair comparison.
  \item $a$ \ljpcriminal\ consists of three subtasks fine, imprisonment with labor, and imprisonment without labor.
  \item $b$ The number of classes under additional quantization scheme. See Table \ref{tbl_ljp_quantization} in Appendix for details.
  \end{tablenotes}
  \end{threeparttable}
  \end{table}
\endgroup

\paragraph{\casename}
Automatic categorization of legal documents or legal questions is often necessary for various downstream tasks.
For instance, for the given factual description of cases from individuals customers one can consider building a lawyer recommendation system based on the predicted case categories.
In this regard, we make \casename\ dataset that consists of 10k facts and case name pairs extracted from first level trials in \precedent. 
In the task, a model is asked to predict case name (case category) for the given facts.
We collect 11,294 examples from 100 most frequent case categories such as ``fraud'' (사기), ``drunk driving'' (도로교통법위반(음주운전)), and ``Violation of the Labor Standards Act'' (근로기준법위반)  where each category includes 100 examples (Table \ref{tbl_tasks}). An example is shown in Appendix \ref{sec: casename example}. 
We also provide \casenamep, a dataset that extends \casename\ by including infrequent case categories. \casenamep\ consists of 31,283 examples with total 603 case categories. The data distribution is shown Fig. \ref{fig_data_dist}a in Appendix.

\paragraph{\statute}
In criminal cases, the ranges of the punishment are determined by corresponding statutes following ``no penalty without a law'' principle. 
Thus predicting the statutes for the given facts is the first step to determine legal judgements.
For this task, we make \statute\ that consists of 3,298 facts and corresponding statutes pairs. The dataset consists of 46 frequent case categories where each class includes 60 examples (Table \ref{tbl_tasks}, Appendix \ref{sec: statute example}). The dataset is extracted from the same precedents used in \casename.
We also release \statutep, a dataset that extends \statute\ by including less frequent case categories. \statutep\ includes 17,730 examples with total 434 case categories and 1,015 statutes. The data distribution is shown Fig. \ref{fig_data_dist}b in Appendix.

\paragraph{\ljpcriminal}
In criminal cases, judges decide the ranges of punishment based on the facts. 
In this regard, we prepare \ljpcriminal\ dataset that consists of 10,500 facts and the corresponding three kinds punishment: (1) fine, (2) imprisonment with labor, and (3) imprisonment without labor (Table \ref{tbl_tasks}). In this task, a model needs to predict the punishment from the given facts. 
We design the task with varying degree of difficulties. At the level 0, a model predicts only the type of punishment if any. At level 1, a model is asked to classify the degree of punishment into three levels null, low, and high. The boundary of low and high is determined in a way that balances the data distribution between two classes (Table. \ref{tbl_ljp_quantization} in Appendix, 3rd and 4th rows). At level 2, we further split the range of punishment into five (fine) and six (imprisonment) empirically based on the data distribution and the certain characteristics of Korean legal system. For instance in the fine prediction subtask, KRW 1,000,000/3,000,000 is selected as a boundary as this is the lower bound where public officials can lose their position if found guilty related to sexual/general crime\footnote{Although there is an official sentencing guideline from the sentencing commission of the Korean government (\url{https://sc.scourt.go.kr/sc/engsc/pdf/SentencingGuidelines2021.pdf}), the guide provides only the ranges of the possible punishments divided by three high levels--reduced, base, weighted--for individual case categories. Finding the quantization boundaries that are more theoretically grounded is currently under study.}. At level 3, a model predicts the exact numbers and the task becomes regression.
The dataset is extracted from the first level criminal trials from \precedent\ with the following case categories: ``indecent act by compulsion'' (강제추행), ``obstruction of performance of official duties'' (공무집행방해), ``bodily injuries from traffic accident'' (교통사고처리특례법위반(치상)), ``drunk driving'' (도로교통법위반(음주운전)), ``fraud'' (사기), ``inflicting bodily injuries'' (상해), and ``violence'' (폭행). Each category includes 1,500 examples. Only trials with a single defendant and a single charge are selected. 

\paragraph{\ljpcivil}
In civil trials, plaintiffs often claim money as a part of compensation and the judges approve a certain portion of the claimed money based on the facts.
For this legal judgement procedure, we built \ljpcivil\ that consists of 4,678 pairs of facts, gist of claim, and the degrees of claim acceptance (Table \ref{tbl_tasks}). 
A model trained on this task can be used to assist judges or legal practitioners on this legal decision making process. 
The claim acceptance degrees are calculated (1) by parsing the claimed money from the gist of claim, (2) extracting the approved money from the ruling section, and (3) dividing the approved money by the claimed money. As a model requires diverse world knowledge for the accurate prediction of the numbers, we simplify this regression task by quantizing the claim acceptance degrees in two levels. 
At level 1, the claim acceptance degrees are partitioned into three categories: rejection, partial approval, and full approval (Table \ref{tbl_ljp_quantization}, 9th row). At level 2, the output ranges are further split into 13 categories with uniform range making the task more close to regression (10th row). 
To build the dataset, it is essential to extract moneys from the gist of claims and rulings which are not trivial to parse as they are often written in a compact form. For instance, instead of an easier expression such as ``A provides \$100 to C, and B provides \$100 to C'', judges would write ``To C, A and B provides \$100 each''. To parse them we train mt5-small \citep{xue2021aclmt5} with a prompt-tuning method \citep{liu2021ptuning,ding2021openprompt} on 160 examples consisting of the pairs of gist of claim (or ruling) and the corresponding parses. The parses consist of the money provider, the money receiver, the amount of money, and the litigation cost. The ruling of criminal cases were parsed similarly. The related technical report is currently under preparation.
We extract \ljpcivil\ from the precedents of following four categories: 929 examples from ``price of indemnification'' (구상금), 745 examples from ``loan'' (대여금), 1,004 examples from ``unfair profits'' (부당이득금), and 2,000 examples from ``lawsuit for damages (etc)'' (손해배상(기)). During the data collection, we do not collect precedents including counterclaims.

\paragraph{\summarization}
For legal practitioners, it is important to survey legal documents efficiently by focusing on the essential parts.
In this regard, we make \summarization\ dataset that consists of 20,000 pairs of precedent from the collection of Supreme Court Decisions Report containing Summary of Decision written by Director of Judicial Research from Supreme Court Library of Korea\footnote{\url{https://www.scourt.go.kr/eng/supreme/decisions/guide.jsp}} (Table \ref{tbl_tasks}).
In this task, a model needs to generate an abstract summary for the given ruling and the reasoning sections of the precedent.
We retrieve the data from Korean Law Information Center\footnote{\url{https://www.law.go.kr}} and collect precedents only when the sum of the number of tokens from the ruling and the reasoning section is less then 1,024\footnote{The length is based on mt5 tokenizer~\citep{xue2021aclmt5}}. The average number of tokens is 527 for the input texts (the ruling and the reasoning sections) and 133 for the ground truth (the summary).
We also provide \summarizationp\ by extending \summarization\ with precedents with longer text making the task more challenging and realistic. In the extended dataset there are a total of 51,114 examples. The average number of tokens in the precedents and the corresponding summaries are 1,516 and 248 respectively. The maximum number of tokens in the input texts and the summaries are 93,420 and 6,536 respectively. The distribution of token length is shown in Fig. \ref{fig_data_dist}c in Appendix.

\section{Experiments}
\subsection{Model training}
All experiments are performed on Nvidia A6000, RTX3090 or RTX6000 hosted in Lambda, iwinv (\texttt{https://www.iwinv.kr/}), or nipa (\texttt{https://www.nipa.kr/eng/index.do}) clouds respectively. We use Transformers library \citet{wolf2020huggin} to fine-tune individual models. All models are trained with a constant learning rate 0.00003--0.0001 with batch size 8--16 using AdamW optimizer \citep{loshchilov2017adamw}. All finetuning experiments with errorbars were repeated three time with different random seeds except \ljpcriminal-Lv0 where the errorbars were estiamted from two independent experiments.
To fine-tune mt5-small, \texttt{google/mt5-small} checkpoint is loaded. For \oursm, we pre-train GPT-2 from scratch using Megatron library \citep{shoeybi2019megatron} using \precedent\ together with Modu and Wiki corpora (Table \ref{tbl_pretraining_staticstics}). During pre-training the examples are randomly sampled from three corpora. We use morpheme-aware byte-level BPE for the tokenization reflecting that Korean is an agglutinative language \citep{kim2021hyperclova}.
We train \oursm-base for 50K steps and \oursm-medium for 100K steps which are well above the minimum number of training steps expected from the power law \citep{ghorbani2022scalinglaws,hoffmann2022chinchilla}.
The model configuration is shown in Table \ref{tbl_pretraining_configuration}.
In the domain adaptation experiments, the KoGPT2 and LCube were trained with \precedent\ corpus for 14 epochs with the batch size 12 using the same objective (maximum likelihood with teacher-forcing). mt5-small was pre-trained with word level span corruption objective for 22 epochs with the batch size 12.

\subsection{Task setting}
We formulate the all tasks as text generation following \citep{raffle2019t5}, not only because this allows an easy extension to other tasks but also it shows its effectiveness even in information extract tasks \citep{hwang2021wyvern,kim2021donut}.

\subsection{Metric}
In \casename, \statute, and \ljpcivil\  tasks, we use string exact match to evaluate the model accuracy. In \ljpcriminal\ task, we calculate $F_1$ of individual fields (fine, imprisonment with labor, imprisonment without labor) as following. When the target field exists in both ground truth (GT) and prediction (1) the case is counted as a true positive if their values are equal and (2) false positive otherwise. (3) When the target field exists only in GT but not in the prediction, the case is counted as a false negative. (4) If the target field is empty in the GT but exists in the prediction it is counted as a false positive. (5) If the field is empty in both GT and the prediction, the case is considered as a true negative. The zero labels in \ljpcriminal\ are counted as nulls.
\begingroup
\setlength{\tabcolsep}{6pt} 
\renewcommand{\arraystretch}{0.8} 

\begin{table}
\centering

\tiny
  \caption{Comparison of various models. For the fair comparison, the \testex\ sets are used which are constructed from the precedents not included in \precedent. In case of \summarization\ task, the summaries are not included in \precedent\ and thus the original test set is used. ``imp.'', ``EM'', ``R1'', ``R2'', and ``RL'' stand for imprisonment, an accuracy from exact match, Rouge-1, Rouge-2, and Rouge-Long scores respectively. ``d.a.'' stands for ``domain adaptation''. ``n$_\text{char}$'' Stands for the average number of characters in the summary. The average n$_\text{char}$ of the GT is 242.
  }
  \label{tbl_baseline}
  \centering
  \begin{threeparttable}

  \begin{tabular}{llllllll}
    \toprule
    \multicolumn{1}{c}{Name} &
    \multicolumn{1}{c}{\makecell{Size}} &
    \multicolumn{1}{c}{\makecell{\casename\\(EM)}} &
    \multicolumn{1}{c}{\makecell{\statute\\ (EM)} } &
    \multicolumn{4}{c}{\summarization} 
    \\
       &  &  & & R1$^*$&R2$^*$&RL$^*$& n$_\text{char}$  \\
    \midrule
      
    KoGPT-2-base   & 125M & $78.5_{\pm 0.3}$  & $85.7_{\pm 0.8}$         
    & $47.2$ & $39.1$ & $45.7$  & $277$
    \\    
    \oursm-base-zero  & 124M  & $79.6 _{\pm 0.6} $         & $ 85.8 _{\pm 0.8}$  
     & $44.9$ & $36.7$ &$43.3$ & $203$ 
     \\    
    mt5-small & 300M & $81.0_{\pm 1.3}$  & $87.2_{\pm 0.3}$ & $56.2$  & $47.8$ & $54.7$ & $299$
    \\
    mt5-large& 1.2B & ${82.9} _{\pm 0.2}$  & $88.1_{\pm 0.5}$    
    & $59.0$ & $50.1$ & $57.6$ & $278$
    \\
    \midrule
    KoGPT-2-base + d.a.   & 125M & $81.9_{ \pm 0.2} $  &  $89.4_{\pm 0.5}$         
     & $49.2$ & $40.9$ & $47.7$  & $273$
    \\            
    \oursm-base-zero + d.a. & 124M & $ 80.4_{\pm 0.2}$  & $ 87.0_{\pm 1.7}$         
    &$46.2$  &$37.6$  & $44.6$   & $200$
    \\
    \oursm-base   & 124M & ${81.1}_{ \pm 0.3}$ & ${87.6}_{\pm 0.5}$  & $46.0$  & $37.7$ & $44.5$ & $195$
    \\
    \oursm-base + d.a. & 124M & $82.7 _{\pm 0.6}$  & $ 89.3_{\pm 0.4}$       
    &$47.8$  &$39.5$  & $46.4$   & $202$  
    \\
    \midrule 
    \oursm-medium  & 354M  & $81.2_{\pm 0.4} $         &  $87.7_{\pm 0.5}$
    &$50.1$ & $41.4$ &$48.6$ & $213$  
    \\
    mt5-small + d.a.  & 300M & $82.2_{\pm 0.2} $  & $ 88.8 _{\pm0.5}$         
     & $56.2$ & $47.7$ & $54.8$ & $291$  
    \\
    \bottomrule
  \end{tabular}
  \begin{tablenotes}[para]
  \item *: Computed at word level.
  \end{tablenotes}
  \end{threeparttable}
  \vspace{-4mm}

\end{table}

\endgroup

\section{Results}
In this section, we show the baseline scores for \casename, \statute, \ljpcriminal, \ljpcivil, and \summarization\ tasks by fine-tuning various language models (Table \ref{tbl_baseline}).
We pre-train a decoder-only Transformer language model based on GPT-2 architecture. Decoder only models are known to show strong zero-shot performance compared to encoder-decoder language models and can be easily converted to non-causal decoder by additional pre-training for fine-tuning tasks \citep{scao2022what_lm_limited_budget,wang2022what_language_decoder_only}. 
We prepare \oursm-base, \oursm-medium (Table \ref{tbl_pretraining_configuration}), and LCube-base-zero, a twin model of LCube-base that is pre-traiend without using \precedent. We compare our models to KoGPT-2, a Korean GPT-2 made from SKT\footnote{\url{https://github.com/SKT-AI/KoGPT-2}}, mt5-small, and mt5-large \citep{xue2021aclmt5} with or without domain adaptation.

\paragraph{Domain specific corpus is critical in the classification and the summarization tasks}
In \casename\ classification task, \oursm-base shows a superior performance compared to \oursm-base-zero (+1.5\%, 2nd vs 7th rows in Table \ref{tbl_baseline}) and KoGPT2 (+2.6\%, 1st vs 7th rows). Similarly, in \statute\ classificatoin task, \oursm-base shows +1.8\% and +1.9\% compared to \oursm-base-zero and KoGPT2 respectively. This clearly shows the importance and effectiveness of using domain specific corpus (\precedent) during the pre-training.

Next we perform domain adaptation experiments by further pre-training KoGPT2, \oursm, and mt5-small with \precedent\ only.
All models shows higher scores upon the domain adapation in (\casename, \statute, \summarization) tasks; KoGPT2 shows (+3.4\%, +3.7\%, +2.0 RL, 1st vs 5th rows); \oursm-base-zero shows (+0.8\%, +1.2\%, +1.3 RL, 2nd vs 6th rows); \oursm-base shows (+1.6\%, +1.7\%, +1.9 RL, 7th vs 8th rows); mt5-small (+1.2\%, +1.6\%, +0.1 RL, 3rd vs final rows).
Notably, \oursm-base shows a comparable performance with mt5-large, a competitive encoder-decoder language model with 860\% more parameters (4th vs 8th rows). 

Interestingly, in \summarization\ task, \oursm\ does not seem to have an advantage over other models (columns 5--7). Further analysis reveals, \oursm\ has a tendency to generate $\sim$40\% fewer tokens compared to other models (2nd, 6--9th rows in the final column) which decreases the rouge scores. The performance of language models on downstream tasks is recently found to be dependent on pre-training corpus \citep{shin2022corpora_effect_hyperclova} which may account for the result\footnote{Although KoGPT-2 and \oursm\ both were pre-trained using Modu and Korean Wikipedia corpora, KoGPT2 uses additional News, Petitions to the Blue House, and other unknown internal corpus.}.
The superior performance of mt5 in \summarization\ task over other models may be originated from their architectural difference as encoder-decoder language models show stronger performance over decoder-only models on the tasks where understanding long context is necessary like a summarization task \citep{soltan2022arXiv_alexatm}. 
On \casenamep\ task where the domain of the task extended by including infrequent case categories, \oursm-base-zero achieves 75.6\% exact match score (EM) whereas \oursm-base achieves 78.9\% EM. Similarly on \statutep\ task, \oursm-base-zero achieves 82.5\% EM whereas \oursm-base shows  84.9\% EM. Again, both results show the importance and effectiveness of \precedent.

\begingroup
\setlength{\tabcolsep}{2pt} 
\renewcommand{\arraystretch}{0.8} 

\begin{table}
\centering

\scriptsize
  \caption{Comparison of various models on \ljpcivil\ task. "EM" stands for exact match.
  }
  \label{tbl_baseline_ljp_civil}
  \centering
\begin{threeparttable}
  \begin{tabular}{llll}
    \toprule
    \multicolumn{1}{c}{Name} &
    \multicolumn{1}{c}{\makecell{size}} &
    \multicolumn{1}{c}{\makecell{\ljpcivil\\Lv 1\\ (EM)} } &
    \multicolumn{1}{c}{\makecell{\ljpcivil\\Lv 2\\(EM)}}  
    \\
    \midrule

    KoGPT-2-base   & 125M       
      & $66.0_{\pm0.5}$ & $58.4_{\pm 0.1}$
     \\    
    \oursm-base-zero  & 124M   & $65.4 _{\pm 0.9} $  & $58.6_{\pm 0.1}$
    \\   
    mt5-small (512 gen)  & 300M    & $68.9_{\pm 0.8}$ & $57.7_{\pm0.6 }$
    \\
    mt5-large (512 g)& 1.2B 
     & $69.0 $ & $58.6$
    \\    
    \midrule 
    
    KoGPT-2-base + d.a.  & 125M & $64.7 _{\pm 1.1}$& ${55.2_{\pm 0.5}}$
    \\        
    \oursm-base-zero + d.a. & 124M      
    & $63.9_{\pm 1.5} $   & $52.7_{\pm 0.4}$
    \\
    \oursm-base   & 124M   &  $ {67.6}_{ \pm 1.3}$  &   $58.8_{\pm  0.0}$
    \\

    \oursm-base + d.a. & 124M        
      & $60.9_{\pm 1.1} $   &  $52.6_{\pm0.4 }$
    \\    
    \midrule
    \oursm-medium  & 354M    & $68.9_{\pm 1.6} $   &   $58.2_{\pm 0.1}$
    \\
    \oursm-medium + d.a. & 354M        
    &  $68.5_{\pm 1.2} $   &  $55.7_{\pm0.6 } $
    \\    
    mt5-small + d.a.  & 300M 
    &  $69.1_{\pm0.1} $ & $58.8_{\pm 0.2}$
    \\        
       
    \midrule
    \oursm-base - facts  & 124M & $58.5$ &  -  
    \\
    Most frequent label  & - &$56.6$ &  $57.3$ 
    \\

    \bottomrule
  \end{tabular}
  \begin{tablenotes}[para]
  
  \end{tablenotes}
  \end{threeparttable}
  \vspace{-4mm}

\end{table}

\endgroup

\paragraph{Domain adaptation is not helpful on legal judgement prediction tasks} \label{sec: ljp_civil}
In \ljpcivil\ task, a model needs to predict the degrees of claim acceptance for the given facts and the gist of claim. The degrees of claim acceptance indicates the ratio between the approved money by the judge and the claimed money from the plaintiffs. To simplify the task, we first convert the task as a classification task having three labels: (1) label 0 for the case where the claim from plaintiffs is completely rejected, (2) label 1 when the claimed money is partially approved, and (3) label 2 when the claim is fully approved (9th row, Table \ref{tbl_ljp_quantization} in Appendix).
\oursm\ achieves higher accuracy compared to \oursm-base-zero (+2.2\%) and KoGPT2 (+1.6\%) (3rd column, 1st, 2nd rows vs 7th row). 
It is noticeable that the domain adaptation always decreases the accuracies of the decoder-only models on the contrary to the case of \casename, \statute, and \summarization\ tasks; -1.3\% for KoGPT2 (1st vs 5th rows); -1.5\% for \oursm-base-zero (2nd vs 6th rows); -6.7\% in \oursm-base (7th vs 8th rows); -0.4\% for \oursm-medium (9th vs 10th rows). 
The accuracy of mt5-small also does not increase under the domain adaptation (3th vs 11th rows). 
This is in line with the result from the previous study showing that for more difficult tasks, pre-training from scratch is more helpful than domain adaptation \citep{chalkidis2020legalbert}. 
When only the gist of the claims are used as the input to the model excluding the facts, the accuracy drops by 9.1\% (7th vs 12th rows), highlighting the importance of the facts for the judgement prediction task. The test set consists of 56.6\% rejection, 41.4\% partial accept, 2.0\% full accept. This indicates that without the facts, the model performance is close to a dummy baseline that selects the most frequent label (i.e. rejecting every claim).

To check whether the result is not from an artifact of the quantization of the claim acceptance ratio, we perform additional experiments with more fine-grained labels (Table \ref{tbl_ljp_quantization} in Appendix, final row). Under this new setting, the claim acceptance ratio is partitioned with 0.1 interval mimicking the regression setting (when the range is quantized uniformly and their interval becomes infinitesimal, a classification task becomes a regression task \citep{bishop:2006:PRML}). 
Again we found the domain adaptation is not helpful; -3.2\% for KoGPT2 (1st vs 5th rows, final column); -4.1\% for \oursm-base-zero (2nd vs 6th rows); -6.2\% in \oursm-base (7th vs 8th rows); -2.5\% for \oursm-medium (9th vs 10th rows); +0.2\% for mt5-small (4th vs 11th column). 
Similarly, on \ljpcriminal\ task, the domain adaptation does not show clear improvement on the model performance (see next section).

\begingroup
\setlength{\tabcolsep}{2pt} 
\renewcommand{\arraystretch}{0.8} 

\begin{table}
\centering

\tiny
  \caption{Comparison of various models on \ljpcriminal\ task.   }
  \label{tbl_baseline_ljp_criminal}
  \centering
\begin{threeparttable}
  \begin{tabular}{ll|lll|lll|lll}
    \toprule
    \multicolumn{1}{c}{Name} &
    \multicolumn{1}{c}{\makecell{size}} &
    \multicolumn{3}{c}{\makecell{Lv 0}} &
    \multicolumn{3}{c}{\makecell{Lv 1}} &
    \multicolumn{3}{c}{\makecell{Lv 2}} 
    \\
      &  
      & fine & \makecell{imp. \\w/ labor} & \makecell{imp.  \\w/o labor}
      & fine & \makecell{imp. \\w/ labor} & \makecell{imp.  \\w/o labor} 
      & fine & \makecell{imp. \\w/ labor} & \makecell{imp.  \\w/o labor}
      \\
    \midrule
    KoGPT-2-base   & 125M
    & $ 69.0_{ \pm 0.7}$  &$ {83.2}_{ \pm 0.3}$ &$ {82.6}_{ \pm 0.2} $   
    & $ 61.0_{ \pm 0.3}$  &$ {72.7}_{ \pm 0.0}$ &$ {59.3}_{ \pm 1.0} $ 
    &$ {49.9}_{\pm 1.7}$ & $ 67.5_{\pm 1.1}$ & $69.2_{ \pm 1.6}$
    \\    
    \oursm-base-zero & 124M  
    & $ 69.1_{\pm 0.3}$ & $82.7_{\pm 0.1}  $ & $81.9_{\pm 1.0} $
    & $ 62.5_{\pm 0.2}$ & $71.1_{\pm 1.3}  $ & $55.6_{\pm 0.1} $
    & $ 49.8$ & $ 65.4$ & $70.1 $ 
    \\    
    mt5-small & 300M 
    & $ 69.8_{\pm 0.1}$ & $83.4_{\pm 0.5}  $ & $83.3_{\pm 0.7} $
    & $ 62.0_{\pm 0.3}$ & $71.0_{\pm 1.1}  $ & $55.9_{\pm 0.9} $
    &  $49.1_{\pm 1.3}$ & $66.6_{\pm 0.6}$  & $69.8_{\pm 1.0}$ 
    \\
    mt5-large (512 g)& 1.2B 
    & $ 69.6_{\pm 0.3}$ & $83.5_{\pm 0.6}  $ & $84.0_{\pm 0.5} $
    & $ 61.7_{\pm 0.3}$ & $72.4_{\pm 0.5}  $ & $58.1_{\pm 0.4} $
    & $46.9 $ & $ 68.7$ & $69.5 $ 
    \\
    \midrule 
    KoGPT-2-base + d.a.  & 125M 
    & $ 68.2_{ \pm 0.4}$  &$ {82.6}_{ \pm 0.8}$ &$ {83.0}_{ \pm 0.4} $ 
    & $ 60.9_{ \pm 0.1}$  &$ {71.5}_{ \pm 0.0}$ &$ {55.6}_{ \pm 0.6} $ 
    & $ 51.1_{\pm 0.9}$ & $65.7_{\pm 2.5} $ & $68.8_{\pm 1.0} $ 
    \\
    \oursm-base   & 124M 
    & $ 67.2_{ \pm 2.6}$  &$ {82.2}_{ \pm 0.6}$ &$ {81.5}_{ \pm 0.2} $ 
    & $ 60.1_{ \pm 1.7}$  &$ {71.0}_{ \pm 0.3}$ &$ {55.2}_{ \pm 0.3} $ 
    & $ 46.4_{ \pm 2.8}$  &$ {69.3}_{ \pm 0.3}$ &$ {70.3}_{ \pm 0.7} $ 
    \\
    \oursm-base + d.a. & 124M 
    & $ 66.8_{ \pm 1.7}$  &$ {80.7}_{ \pm 0.4}$ &$ {81.6}_{ \pm 1.2} $ 
    & $ 59.4_{ \pm 2.1}$  &$ {69.3}_{ \pm 0.4}$ &$ {55.9}_{ \pm 3.5} $ 
    &$ 48.1_{\pm 1.2}$ & $67.4_{\pm 1.5} $ & $69.9_{\pm 1.1} $
    \\
    \midrule
    \oursm-medium  & 354M  
    & $ 69.2_{ \pm 1.9}$  &$ {81.9}_{ \pm 1.5}$ &$ {82.7}_{ \pm 0.3} $ 
    & $ 61.2_{ \pm 2.1}$  &$ {70.8}_{ \pm 1.8}$ &$ {57.6}_{ \pm 0.7} $ 
    & $ 49.3_{\pm 0.8}$ & $67.3_{\pm 1.9} $ & $68.1_{\pm 1.0} $
    \\
    \oursm-medium + d.a. & 354M 
    &  $ 69.6_{\pm 1.4}$ & $83.8_{\pm 0.1} $ & $84.0_{\pm 1.1} $
    & $ 61.4_{\pm 0.4}$ & $72.8_{\pm 0.6} $ & $53.2_{\pm 0.3} $
    &$ 48.6_{\pm 2.0}$ & $69.3_{\pm 0.1} $ & $69.8_{\pm 0.5} $
    \\

    mt5-small + d.a.  & 300M 
    & $ 67.9_{\pm 0.4}$ & $82.1_{\pm 1.1}  $ & $80.8_{\pm 0.6} $
    & $ 61.3_{\pm 1.1}$ & $71.2_{\pm 1.1}  $ & $54.0_{\pm 3.3} $
    & $ 51.8_{\pm 0.7}$ & $68.9_{\pm 0.3} $ & $70.3_{\pm 0.1} $
    \\
    \midrule
    Most frequent label & - 
    & $56.6$ & $67.9$ & $15.5$
    & $34.3$ & $45.2$ & $8.8$
    & $30.1$ & $42.8$  & $11.9$ 
    \\
    \oursm-base - facts + casename & 124M 
    & $63.5$ & $75.0$ &$74.6$
    & $55.3$ & $65.6$ & $49.4$
    & $42.0$ & $64.0$ &$61.4$
    \\
    \oursm-base + reason $^\dagger$ & 124M 
    & $77.5$  & $87.2$ & $78.7$
    & $69.1$ & $76.2$ & $53.0$
    & $60.5$  & $74.$6  & $72.7$ 
    \\
    \oursm-medium + reason $^\dagger$ & 124M 
    & $81.8$  & $87.5$ & $86.6$
    & $74.0$ & $77.3$ & $63.1$
    & $60.4 $ & $76.1 $ & $ 76.1$
    \\
    \midrule 
    Null ratio& - 
    & 0.605 & 0.486 & 0.916
    & 0.605 & 0.486 & 0.916
    & 0.605 & 0.486 & 0.916 
    \\  
    Top non-null label ratio & - 
    & 0.395 & 0.514 & 0.084
    & 0.207 & 0.292  & 0.046 
    & 0.188 & 0.272 & 0.063
    \\
    Top non-null label ratio (w/o null) & - 
    & 1.0 & 1.0 & 1.0
    & 0.523 & 0.568 & 0.551
    & 0.474 & 0.528  & 0.744 
    \\

    \bottomrule
  \end{tabular}
  \begin{tablenotes}[para]
  $\dagger$ The reason for the sentencing. 
  \end{tablenotes}
  \end{threeparttable}

\end{table}
\endgroup

\paragraph{Legal judgement prediction is challenging} \label{sec: ljp_criminal}
In \ljpcriminal\ task, a model needs to predict the fine amount, the range of imprisonment with labor or imprisonment without labor for the given facts. Given the difficulty of the task, we approach the task with three different levels of difficulty: (1) level 0 where a model only needs to predict the type of punishment if any, (2) level 1 where a model predicts the punishment in three degrees (null, low, high), and (3) level 2 where the ranges of punishments are further partitioned into five (fine) and six (imprisonment) labels.
Not surprisingly, the scores decrease at higher level (Table \ref{tbl_baseline_ljp_criminal}). Except at level 0, the models show much lower performance compared to \casename\ and \statute.
To confirm the importance of the facts for the prediction, we replace the facts in the inputs to the case names. The $F_1$ scores for fine, imprisonment with labor, and imprisonment without labor prediction tasks drop by  -3.7\%, -7.2\%, and -6.9\% at the level 0,  -4.8\%, -5.4\%, and -5.8\% at the level 1, and -4.4\%, -5.3\%, and -8.9\% at the level 2 (6th vs 12th rows). This clearly indicates that the facts include importance information for the judgement prediction. When the predicted labels for the individual subtasks are replaced by the most frequent non-null labels for all cases, the $F_1$ scores decrease further (11th row). The relatively higher scores observed when the case name are used as the inputs may be originated from the fact that the cases with the same category will likely to have relatively similar punishments ranges. 
As another control experiment, we add ``the reason for the sentencing'' section of the precedents to the inputs. In this section, the judges describe various aspects that can affect the final results such as upper and lower bounds of punishment ranges described in the related statutes, ages and attitude of defendants, victim's opinions etc. This can be considered as an incomplete oracle. Upon the addition of the oracle, there is dramatic increase in $F_1$'s (6th vs 13th rows, and 8th vs 14th rows).
Interestingly, unlike other tasks, we could not find the single superior model even with additional experiments under regression setting (Table \ref{tbl_baseline_ljp_criminal_regression} in Appendix). 

\section{Conclusion}
With the rapid advancement of deep learning, how artificial intelligence (AI) can be applied in legal domain is fundamentally changing. In response to this, we release the first large-scale Korean legal AI benchmark \ours\ and Korean legal language model \oursm. \ours\ consists of one legal corpus, two text classification tasks, two challenging legal judgement predictions tasks, and one precedent summarization task. Experiments on various language models highlight the importance of pre-training language models on domain-specific corpus in large scale. In the future, we plan to extend \ours\ to include more diverse legal language understanding tasks such as court opinion generation.

\section*{Limitation and Future Works}
In this work, we only consider the precedents from the first level courts. In the trials from the second level courts, the legal reasoning can become more complex due to the combination of different appeals from each party. We also do not use claims from plaintiffs and defendants as an additional input to a model although they include important information for various legal decisions. 
Without a high-performance parser, it is difficult to separate the claims from reasoning sections without errors as they are loosely separated from the facts and the reasoning. The claims will be integrated to the future version of \ours.
Our datasets also currently do not consider many important legal applications of AI such as a legal information retrieval task. This will be integrated in \ours\ in the future.

\section*{Ethical Considerations}
The precedents from criminal courts often include detailed descriptions about crimes. Thus one can obtain the essential information for the illegal acts by reading the related precedents.
Nevertheless, laws alone cannot comprehensively cover complex social phenomena and precedents often fill the empty spaces of the laws. Thus, we believe, enabling an easy access to precedents can bring more social benefits than harms.

The models trained on the legal judgement prediction (LJP) datasets can be used for automatic legal decisions and like any data-driven approaches, the models can predict biased results depending on the data statistics of the training set. To minimize the risk, we have used only the precedents redacted by the Korean government following the official protocol (Section \ref{sec: korean precedent}). Also, we have fully open-sourced the dataset, trained model, and training code so that anyone can access and investigate the dataset and the method without any restriction.

Nevertheless as the population of precedents is already biased reflecting the current status of the society (for example, the judicial yearbook published in 2021 reveals the 86.8\% defendants in 1st criminal trials were male\footnote{\url{https://www.scourt.go.kr/img/pub/jur_2021_Book6.pdf}}, and as we randomly sampled the population during the construction of \ours\ whenever possible, the use of the released dataset and the model without explicitly understanding the potential issue from the dataset bias can make negative social impacts.

To concretely address this issue, we performed a case study about gender bias in the criminal precedents (race, language, and regions of litigants are not considered as our datasets consist of Korean cases only). The age is also not considered at this time as judges can reduce or increase the sentences based on the ages of litigants following the official sentencing guidelines.

We first extracted the victims' gender information from the facts using regular expressions. Among five criminal case categories in \ljpcriminal\ dataset, only “indecent act by compulsion” (강제추행) shows strong bias in victims' gender (1,451 females, 74 males). We examined how the prediction of LCube changes when the victims' genders are exchanged (as only 0.46\% cases include the gender of the defendant in the facts, we change the gender of the victims). Among 131 test examples, the prediction changed in 11 examples. Further analysis revealed in 10 examples (7.6\%) the outcome sentences decreased when the victims' genders were changed from female to male. Although this is relatively small (91.5\% of the predictions remain identical), the result shows the potential risk and the limitation of automatic legal decision making based on our dataset and the released model. Finally we would like to remark that there is also a social benefit in democratizing the data.

\begin{ack}
We thank Gene Lee for his helpful comments and advisory (as a licensed lawyer in Korea) on the legal aspects of this work and carefully proofreading the paper. We also thank the anonymous reviewers for their critical comments.

\end{ack}



{
\small
\bibliography{legal_ai}
\bibliographystyle{acl_natbib}
}

\newpage
\section*{Checklist}

Checklist section heading above along with the questions/answers below.

\begin{enumerate}

\item For all authors...
\begin{enumerate}
  \item Do the main claims made in the abstract and introduction accurately reflect the paper's contributions and scope?
    \answerYes{}
  \item Did you describe the limitations of your work?
    \answerYes{}
  \item Did you discuss any potential negative societal impacts of your work?
    \answerYes{}
  \item Have you read the ethics review guidelines and ensured that your paper conforms to them?
    \answerYes{}
\end{enumerate}

\item If you are including theoretical results...
\begin{enumerate}
  \item Did you state the full set of assumptions of all theoretical results?
    \answerNA{}
	\item Did you include complete proofs of all theoretical results?
    \answerNA{}
\end{enumerate}

\item If you ran experiments (e.g. for benchmarks)...
\begin{enumerate}
  \item Did you include the code, data, and instructions needed to reproduce the main experimental results (either in the supplemental material or as a URL)?
    \answerYes{}
  \item Did you specify all the training details (e.g., data splits, hyperparameters, how they were chosen)?
    \answerYes{}
	\item Did you report error bars (e.g., with respect to the random seed after running experiments multiple times)?
    \answerYes{} 
	\item Did you include the total amount of compute and the type of resources used (e.g., type of GPUs, internal cluster, or cloud provider)?
    \answerYes{}
\end{enumerate}

\item If you are using existing assets (e.g., code, data, models) or curating/releasing new assets...
\begin{enumerate}
  \item If your work uses existing assets, did you cite the creators?
    \answerNA{}
  \item Did you mention the license of the assets?
    \answerNA{}
  \item Did you include any new assets either in the supplemental material or as a URL?
    \answerNA{}{}
  \item Did you discuss whether and how consent was obtained from people whose data you're using/curating?
    \answerYes{} All personal information are anonymized except some special cases where the Korean government decided not to anonymize for the social benefits.
  \item Did you discuss whether the data you are using/curating contains personally identifiable information or offensive content?
    \answerYes{} See previous question.
\end{enumerate}

\item If you used crowdsourcing or conducted research with human subjects...
\begin{enumerate}
  \item Did you include the full text of instructions given to participants and screenshots, if applicable?
    \answerNA{}{}
  \item Did you describe any potential participant risks, with links to Institutional Review Board (IRB) approvals, if applicable?
    \answerNA{}{}
  \item Did you include the estimated hourly wage paid to participants and the total amount spent on participant compensation?
    \answerNA{}
\end{enumerate}

\end{enumerate}


\newpage
\appendix

\section{Appendix}

\subsection{Related Works} \label{sec: related works}
In this section, we compare \ours\ to previous studies. In general, \ours\ differs to other works in that it exclusively treats Korean legal cases especially from the lower courts. 

\cite{chalkidis2019aclLJP} introduces the ECtHR dataset that consists of 11k cases from the European Court of Human Rights. In this task, a model needs to predict a set of related articles for the given facts. The \statute\ dataset also consists of the facts and corresponding statues yet it handles criminal cases from the lower court with more diverse legal domain (5th column in Table \ref{tbl_tasks}).

\cite{niklaus2021nllu_swiss_ljp} releases the Swiss-Judgements-Prediction dataset that consists of 85k multilingual cases--German, French, and Italian--from the Federal Supreme Court of Switzerland. In this task, a model needs to predict ``approval'' or ``dismissal'' regrading the validity of the plaintiff's request for the given facts of the cases. Similarly, in \ljpcivil\ task, a model predicts the claim acceptance degree for the given facts and the gist of claim. However, the dataset differs in that the acceptance degree is quantitatively estimated based on the ratio between the claimed money from the plaintiffs and the approved money by the judges. Also, \ljpcivil\ handles cases from the 1st trials from the district courts of Korea.

\cite{xiao2018cail2018} introduces the CAIL dataset which consists of 2.7m Chinese criminal cases. Given the facts, a model needs to predict corresponding law articles, charges, and prison terms. The dataset is similar to \statute\ (law articles), \casename\ ($\sim$charge), and \ljpcriminal\ ($\sim$prison terms) datasets with following differences; (1) both \statute, and \casename\ include the civil cases; (2) in \ljpcriminal\ task, a model needs to predict three types of punishments, fine, imprisonment with labor, and imprisonment without labor. Additionally, \ljpcriminal\ includes ``the reason for sentencing'' section in which the judges describe various aspects that can affect the final results such as upper and lower bounds of punishment ranges described in the related statutes, ages and attitude of defendants, victim's opinions etc.

\cite{ma2021sigarLJPRealCourtSetting} studies legal judgement prediction over low level data by using plaintiff's claims and court debate as an input. On the contrary, \ljpcivil\ and \ljpcriminal\ relies on the factual description written by judges in precedents. The court debates are not publicly available in Korea.

\cite{chalkidis2022acl_lexglue} introduces a benchmark dataset for legal NLU in English focusing on classification tasks over various legal documents; cases, legislation, contracts, Terms of Service, and holdings in cases. On the other hand, \ours\ present one legal corpus, two classification tasks, two legal judgement tasks (may be considered as classification or regression task in a broader sense), and one summarization task using processed (by ourselves) Korean precedents exclusively.

\cite{chalkidis2022facl_fairlex} investigate legal fairness over four legal judgement datasets with additional attributes such as race, gender, region, language, age etc. As \ours\ handles anonymized Korean cases exclusively, there is almost no data heterogeneity is terms of race, language, and regions and thus the fairness is not investigated as a main topic in this study. On the other hand, we find in \ljpcriminal\ dataset, certain case category such as ``indecent act by compulsion (강제추행)'' shows gender bias. The possible risk and the limitation of \ours\ is discussed in Ethical Considerations.

\subsection{Precedent redaction rule}  \label{sec: redaction_rule}
Data subjected to anonymization are as follows\footnote{\url{https://www.scourt.go.kr/portal/information/finalruling/anony/index.html}, \url{https://glaw.scourt.go.kr/wsjo/gchick/sjo330.do?contId=3202812\#1660965241959}}.
\begin{itemize}
    \item Name and the equivalents: Name, nickname, pen name, ID, and corresponding nouns that point to a specific person are replaced with upper case alphbets A, B, C, etc., without redundancy.
    \item Contact information: Phone number, e-mail address, residential address and corresponding contact data are deleted in the meta information(당사자 단락) and replaced with upper case alphabets in the reasoning(이유 단락)
    \item Financial information: Account number, credit card number, check number and corresponding financial data are replaced with upper case alphabets without redundancy.
    \item Other personally identifible information: Social security number is deleted. car registration number, address of real estate holdings and corresponding personal information are replaced with upper case alphabets without redundancy.
\end{itemize}
In the case of felonies such as serial killing, there are some exceptions where the Korean government can decide not to anonymize for the social benefits\footnote{for example, see \url{https://en.wikipedia.org/wiki/Yoo_Young-chul}, \texttt{https://www.law.go.kr/판례/(2004고합972)}}.

\subsection{Precedent disclosure status} \label{sec: precedent disclosure status}
Opening the precedents public meets the constitutional need to guarantee the right to know and right to a public trial, while allowing the general public to see that the justice system is functioning properly and treating defendants fairly. On the other hand, non-disclosure of certain documents and anonymization are to compromise the stated benefits with one's privacy and honor.

Since the Civil Procedure Act was revised in the December of 2020\footnote{\url{https://law.go.kr/lsInfoP.do?lsiSeq=223439&lsId=&efYd=20230101&chrClsCd=010202&urlMode=lsEfInfoR&viewCls=lsRvsDocInfoR&ancYnChk=0\#}}, all civil court decisions including those not settled and appealed to higher courts will be open to general public from 01.01.2023. In the same context, expanding the scope of disclosure while minimizing the side effects is being discussed in the National Assembly of Korea.

\subsection{The example of Korean precedent} \label{sec: precedent example}
\begin{itemize}
    \item Meta information
        \begin{itemize}
            \item Plaintiffs: A 주식회사
            \item Defendants: 주식회사 B (전동 킥보드 제조 회사)
            \item Case name: 구상금
        \end{itemize}
    \item Gist of claim:
[청구취지]
피고는 원고에게 95,569,454원 및 이에 대하여 2020. 1. 15.부터이 사건 소장 부본 송달일까지는 연 5\%, 그 다음날부터 각 다 갚는 날까지는 연 12\%의 각 비율로 계산한 돈을 지급하라.

    \item Facts: 
[사실관계]
○ 원고는 손해보험업을 영위하는 보험회사로서, 주식회사 C와 생산물배상책임보험(이하 '이 사건 보험계약'이라 한다)을 체결한 보험자이고, 피고는 D 전동킥보드를 제조하여 주식회사 C에 납품하는 업체이다.\\
○ E은 (중략) 전동킥보드를 구입하였고, (중략) 개조를 하였다.\\
○ (중략) 이 사건 전동킥보드를 충전하던 중 화재가 발생하였고 (중략)
...
\item Claim of plaintiffs: 
[원고의 주장]
피고가 제조한 이 사건 전동킥보드 (중략) 전기적 결함으로 인하여 이 사건 화재가 발생한 것이므로, 피고는 (중략) 원고에게 그 보험금 상당액을 지급할 의무가 있다.
\item Reasoning
[판사의 판단]
(중략)제조물 결함으로 인한 배상책임이 인정되기 위하여는, (중략) 해당 제조물의 결함 없이는 (중략) 발생하지 아니한다는 사실이 (중략) 증명되어야 한다. 그런데 (중략) 전동킥보드를 개조함으로써 그 개조 과정의 하자 등으로 인하여 발생한 것일 가능성이 있다는 점에서 이 사건 전동킥보드 배터리 등 자체의 결함으로 인하여 발생한 것이라고 단정하기 어렵고 달리 이를 인정할 만한 충분한 증거가 없다(중략)
\item Ruling:
[판사의 결론]
원고의 청구를 기각한다.
\end{itemize}

\subsection{\ours\ examples} \label{sec: lbox-open-examples}
\subsubsection{\precedent} 
\begin{itemize}
    \item 
  Ruling: 주문 피고인을 징역 6개월에 처한다.다만, 이 판결 확정일로부터 1년간 위 형의 집행을 유예한다.
  \item 
  Reason: 이유 범 죄 사 실 1. 사기 피고인은 2020. 12. 15. 16:00경 경북 칠곡군 B에 있는 피해자 C이 운영하는 ‘D’에서, 마치 정상적으로 대금을 지급할 것처럼 행세하면서 피해자에게 술을 주문하였다. 그러나 사실 피고인은 수중에 충분한 현금이나 신용카드 등 결제 수단을 가지고 있지 않아 정상적으로 대금을 지급할 의사나 능력이 없었다. 그럼에도 피고인은 위와 같이 피해자를 기망하여 이에 속은 피해자로부터 즉석에서 합계 8,000원 상당의 술을 교부받았다. 2. 공무집행방해 피고인은 제1항 기재 일시·장소에서, ‘손님이 술값을 지불하지 않고 있다’는 내용의 112신고를 접수하고 현장에 출동한 칠곡경찰서 E지구대 소속 경찰관 F로부터 술값을 지불하고 귀가할 것을 권유받자, “징역가고 싶은데 무전취식했으니 유치장에 넣어 달라”고 말하면서 순찰차에 타려고 하였다. 이에 경찰관들이 수회 귀가 할 것을 재차 종용하였으나, 피고인은 경찰관들을 향해 “내가 돌로 순찰차를 찍으면 징역갑니까?, 내여경 엉덩이 발로 차면 들어갈 수 있나?”라고 말하고, 이를 제지하는 F의 가슴을 팔꿈치로 수회 밀쳐 폭행하였다. 이로써 피고인은 경찰관의 112신고사건 처리에 관한 정당한 직무집행을 방해하였다. 증거의 요지 1. 피고인의 판시 제1의 사실에 부합하는 법정진술 1. 증인 G, F에 대한 각 증인신문조서 1. 영수증 1. 현장 사진 법령의 적용 1. 범죄사실에 대한 해당법조 및 형의 선택 형법 제347조 제1항, 제136조 제1항, 각 징역형 선택 1. 경합범가중 형법 제37조 전단, 제38조 제1항 제2호, 제50조 1. 집행유예 형법 제62조 제1항 양형의 이유 1. 법률상 처단형의 범위: 징역 1월∼15년 2. 양형기준에 따른 권고형의 범위 가. 제1범죄(사기) [유형의 결정] 사기범죄 > 01. 일반사기 > [제1유형] 1억 원 미만 [특별양형인자] - 감경요소: 미필적 고의로 기망행위를 저지른 경우 또는 기망행위의 정도가 약한 경우, 처벌불원 [권고영역 및 권고형의 범위] 특별감경영역, 징역 1월∼1년 [일반양형인자] 없음 나. 제2범죄(공무집행방해) [유형의 결정] 공무집행방해범죄 > 01. 공무집행방해 > [제1유형] 공무집행방해/직무강요 [특별양형인자] - 감경요소: 폭행·협박·위계의 정도가 경미한 경우 [권고영역 및 권고형의 범위] 감경영역, 징역 1월∼8월 [일반양형인자] - 감경요소: 심신미약(본인 책임 있음) 다. 다수범죄 처리기준에 따른 권고형의 범위: 징역 1월∼1년4월(제1범죄 상한 + 제2범죄 상한의 1/2) 3. 선고형의 결정: 징역 6월에 집행유예 1년 만취상태에서 식당에서 소란을 피웠고, 112신고로 출동한 경찰관이 여러 차례 귀가를 종용하였음에도 이를 거부하고 경찰관의 가슴을 밀친 점 등을 종합하면 죄책을 가볍게 볼 수 없으므로 징역형을 선택하되, 평소 주량보다 훨씬 많은 술을 마신 탓에 제정신을 가누지 못해 저지른 범행으로 보이고 폭행 정도가 매우 경미한 점, 피고인이 술이 깬 후 자신의 경솔한 언동을 깊이 반성하면서 재범하지 않기 위해 정신건강의학과의 치료 및 상담을 받고 있는 점, 식당 업주에게 피해를 변상하여 용서를 받은 점, 피고인의 나이와 가족관계 등의 사정을 참작하여 형의 집행을 유예하고, 범행 경위와 범행 후 피고인의 태도 등에 비추어 볼 때 재범의 위험성은 그다지 우려하지 않아도 될 것으로 보여 보호관찰 등 부수처분은 부과하지 않음. 이상의 이유로 주문과 같이 판결한다.

\end{itemize}

\subsubsection{\casename} \label{sec: casename example}
\begin{itemize}
    \item Facts (input): 질병관리청장, 시·도지사 또는 시장·군수·구청장은 제1급 감염병이 발생한 경우 감염병의 전파방지 및 예방을 위하여 감염병의심자를 적당한 장소에 일정한 기간 격리시키는 조치를 하여야 하고, 그 격리조치를 받은 사람은 이를 위반하여서는 아니 된다. 피고인은 해외에서 국내로 입국하였음을 이유로 2021. 4. 21.경 감염병의심자로 분류되었고, 같은 날 창녕군수로부터 ‘2021. 4. 21.부터 2021. 5. 5. 12:00경까지 피고인의 주거지인 경남 창녕군 B에서 격리해야 한다’는 내용의 자가격리 통지서를 수령하였다. 1. 2021. 4. 27.자 범행 그럼에도 불구하고 피고인은 2021. 4. 27. 11:20경에서 같은 날 11:59경까지 사이에 위 격리장소를 무단으로 이탈하여 자신의 승용차를 이용하여 경남 창녕군 C에 있는 ‘D’ 식당에 다녀오는 등 자가격리 조치를 위반하였다. 2. 2021. 5. 3.자 범행 피고인은 2021. 5. 3. 10:00경에서 같은 날 11:35경까지 사이에 위 격리장소를 무단으로 이탈하여 자신의 승용차를 이용하여 불상의 장소를 다녀오는 등 자가격리 조치를 위반하였다.
    \item Case name (ground truth): 감염병의예방및관리에관한법률위반

\end{itemize}

\subsubsection{\statute} \label{sec: statute example}
\begin{itemize}
    \item Facts (input): 1. 사문서위조 피고인은 2014. 5. 10.경 서울 송파구 또는 하남시 이하 알 수 없는 장소에서 영수증문구용지에 검정색 볼펜을 사용하여 수신인란에 ‘A’, 일금란에 ‘오천오백육십만원정’, 내역 란에 ‘2010가합7485사건의 합의금 및 피해 보상금 완결조’, 발행일란에 ‘2014년 5월 10일’이라고 기재한 뒤, 발행인 옆에 피고인이 임의로 만들었던 B의 도장을 찍었다. 이로써 피고인은 행사할 목적으로 사실증명에 관한 사문서인 B 명의의 영수증 1장을 위조하였다. 2. 위조사문서행사 피고인은 2014. 10. 16.경 하남시 이하 알 수 없는 장소에서 피고인이 B에 대한 채무를 모두 변제하였기 때문에 B가 C회사에 채권을 양도한 것을 인정할 수 없다는 취지의 내용증명원과 함께 위와 같이 위조한 영수증 사본을 마치 진정하게 성립한 문서인 것처럼 B에게 우편으로 보냈다. 이로써 피고인은 위조한 사문서를 행사하였다.
    \item Statutes (ground truth): 형법 제231조, 형법 제234조
\end{itemize}

\subsubsection{\ljpcriminal} \label{sec: ljpcriminal example}
\begin{itemize}
    \item Facts (input): 피고인은 2016.4. 6.경 지인으로부터 피해자 B을 소개받아 알게 된 사이이다. 피고인은 2016. 4. 8.경 장소불상지에서 피해자에게 전화하여 `일수 빚을 갚아야 하니 돈을 빌려주면 반드시 변제하겠다'고 거짓말하였다. 그러나 피고인은 특별한 재산이 없고 1,400만 원 이상 다액의 일수 및 대출금 채무가 있었던 상황으로 피해자로부터 금원을 빌리더라도 이를 변제할 의사나 능력이 없었다. 그럼에도 불구하고 피고인은 위와 같이 피해자를 기망하여 이에 속은 피해자로부터 2016. 4. 8. 경 부산 부산진구 C에 있는 D은행 가야동지점 인근에서 현금 200만 원을 교부받은 것을 비롯하여, 그 무렵부터 2016. 10. 11. 경까지 별지 범죄일람표 기재와 같이 피해자로부터 총 17회에 걸쳐 차용금 명목으로 2,435만 원을 교부받았다.
    \item Punishment (ground truth): 징역 3월
    \item Punishment (quantized ground truth): 벌금0, 징역1, 금고0
    \item Reason for sentencing (optional input): 양형의 이유. 1. 양형기준에 따른 권고형의 범위 [유형의 결정] 사기범죄 > 01. 일반사기 > [제1유형] 1억 원 미만 [특별양형인자] 감경요소: 처벌불원 또는 상당부분 피해 회복된 경우 [권고영역 및 권고형의 범위] 감경영역, 징역 1월~1년 2. 선고형의 결정: 피고인이 진지하게 반성하는 점, 피해회복 후 피해자와 원만히 합의한 점, 별다른 처벌전력이 없는 점, 그밖에 피고인의 연령, 성행, 환경, 범죄 후의 정황 등을 고려해서 주문과 같이 형을 정한다.
    \item Ruling: 피고인을 징역 3월에 처한다. 다만 이 판결 확정일로부터 1년간 위 형의 집행을 유예한다.
\end{itemize}

\subsubsection{\ljpcivil} \label{sec: ljpcivil example}
\begin{itemize}
    \item Facts (input 1): 가. 원고는 (차량번호 1 생략) 포터 화물차(이하 ‘원고 차량'이라 한다)에 관하여 그 소유자인 B와 자동차종합보험계약을 체결한 보험자고, 피고는 사고 장소인 도로의 관리주체다.
나. B는 2019. 12. 13. 15:40경 원고 차량을 운전하여 경북 성주군 금수면 명천리 도로 부근에서 전방 및 좌우를 살펴 조향 및 제동장치를 정확하게 조작하며 안전하게 운전하여야 할 주의의무를 위반하여 원고 차량 기준 우측에서 좌측으로 진행하는 오토바 이를 충격하게 되었다(이하 ‘이 사건 사고'라 한다).
다. 이 사건 사고로 인하여 오토바이 운전자와 동승자가(이하 ‘망인들'이라 한다) 사망하고, 오토바이가 파손되었으며, 원고는 오토바이 망인들의 사망보험금 및 오토바이 수리비로 합계 333,931,050원을 지급하였다.
    \item Gist of claim (input 2) 청구취지. 피고는 원고에게 66,786,210원과 이에 대하여 2020. 3. 19.부터 이 사건 소장 부본 송달일까지는 연 5\%의, 그 다음날부터 다 갚는 날까지는 연 12\%의 각 비율로 계산한 돈을 지급하라.
    \item The degrees of claim acceptance level (ground truth): 0.5 ($=33393105 / 66786210 $)
    \item The degrees of claim acceptance level (quantized ground truth): 부분 인정
    \item Ruling: 1. 피고는 원고에게 33,393,105원과 이에 대하여 2020. 3. 19.부터 2022. 1. 13.까지는 연 5\%의, 그 다음날부터 다 갚는 날까지는 연 12\%의 각 비율로 계산한 돈을 지급하라.
2. 원고의 나머지 청구를 기각한다.
3. 소송비용은 50\%는 원고가, 나머지는 피고가 각 부담한다.
4. 제1항은 가집행할 수 있다.

\end{itemize}

\subsection{Data engineering pipeline} \label{sec: pipeline}
Here we provide the additional details of the precedent engineering pipeline. 
Layout-classifier was prepared by training custom ResNet using 300k training examples consisting of document image and label pairs.
Layout-parser is based on Mask-R-CNN and was trained on 162k examples consisting of document image and coordinates of the figures/tables pairs using detectron2 library \citep{wu2019detectron2}. The initial checkpoint was prepared by loading pre-trained model on PubLayNet dataset \citep{Zhong2019publaynet}.
For Post-OCR Text corrector, we use character level transformers trained on 137 k examples that consist of OCR outputs and manually corrected text pairs.
The initial checkpoint was loaded from KorCharElectra\footnote{\url{https://github.com/monologg/KoCharELECTRA}}. 
The transformer generates tags indicating whether to "keep", "delete", "replace", or "insert" for individual characters. When the generated tags are either "replace" or "insert", the transformer generates an additional new character. 
All training sets were prepared via our own data labeling platform LWorks (https://lworks.kr/work) where we employ $\sim$100 part-time annotators. 

The precision and recalls of individual ML modules on the static test sets are as follows;
Layout-classifiers, P 99.94\%, R 82.0\%;
Layout-parser, P 98.6\%, R 97.1\% with IOU threshold 0.5;
Post OCR Corrector, P 91.1\%, R 74.6\%.

The overall automation efficiency on the deployed system is $\sim$80\%  for PDF-type precedents and $\sim$53\% for non-readable type precedents. The remaining precedents are manually monitored and corrected via the LWorks platform.
We are currently preparing a separate technical report explaining the further details of the method.

\subsection{Data statistics}

\begin{figure}[h]
\centering
\includegraphics[width=0.9\textwidth]{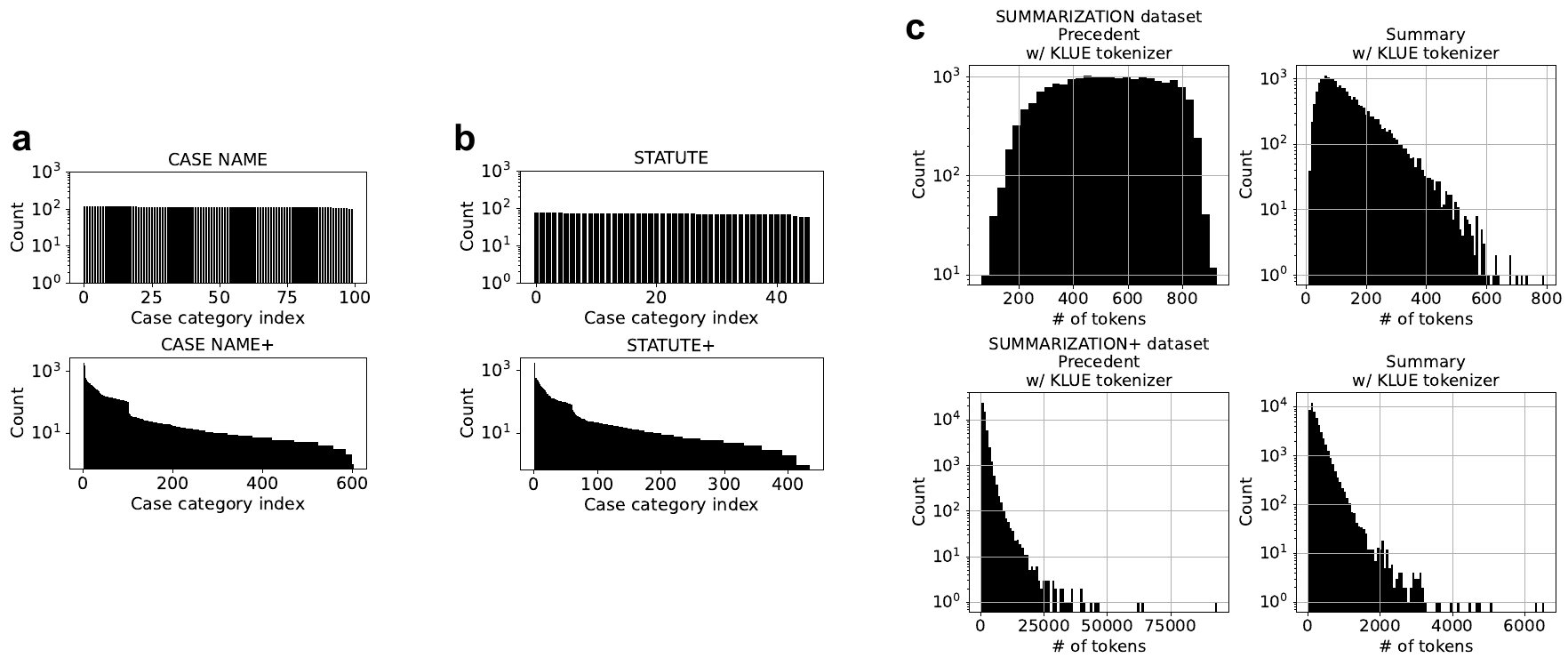}
\caption{Data distribution histograms.} 
\label{fig_data_dist}
\vspace{-4mm}

\end{figure}
\subsubsection{\ljpcriminal\ statistics}
\begingroup
\setlength{\tabcolsep}{2pt} 
\renewcommand{\arraystretch}{0.8} 

\begin{table}[h]
\scriptsize
  \caption{Statistics of \ljpcriminal\ dataset}
  \label{tbl_ljp_criminal_staticstics}
  \centering
  \begin{tabular}{l|ccc}
    \toprule
    
    Case category & fine & imprisonment w/ labor  & imprisonment w/o labor  \\
    \midrule
    
    ``indecent act by compulsion'' (강제추행) & 857 & 756 & 0  \\
    ``obstruction of performance of official duties'' (공무집행방해)& 421  & 1,210 & 0 \\
    ``bodily injuries from traffic accident'' (교통사고처리특례법위반(치상)) & 656 & 2  & 955 \\
    ``drunk driving'' (도로교통법위반(음주운전)) & 228 & 1,414 & 0 \\
    ``fraud'' (사기)  &300  &1,311  &0 \\
    ``inflicting bodily injuries'' (상해) &792  & 811 &0 \\
    ``violence'' (폭행)& 1,286 & 276 & 0\\
    \midrule
    total & 4,540 &5,780 & 955\\

    \bottomrule
  \end{tabular}
\end{table}
\endgroup

\subsection{The domain quantization scheme of two LJP tasks}
\begingroup
\setlength{\tabcolsep}{2pt} 
\renewcommand{\arraystretch}{0.4} 

\begin{table}[!h]
 \centering

\scriptsize
  \caption{Class labels for legal judgement prediction tasks.}
  \label{tbl_ljp_quantization}
  {
  \centering
  \begin{tabular}{lcccccc}
    \toprule
    Tasks   & \# of classes  & subtask & unit & ranges  & label & number \\
    \midrule
    \midrule
    \ljpcriminal-Lv0 &2 &fine & KRW 1000 & \makecell[l]{0 (null label) \\ $0< \cdot$}&\makecell[c]{0\\1} & \makecell[c]{6,888\\4,540}\\
    \midrule
    \ljpcriminal-Lv0 &\makecell{2\\2} &\makecell[l]{imprisonment with labor,\\imprisonment without labor} & month& \makecell[l]{0 (null label) \\ $0 < \cdot $ } &\makecell[c]{0\\1} & \makecell[c]{16,121\\6,735}   \\
    
    \midrule
    \midrule
    \ljpcriminal-Lv1 &3 &fine & KRW 1000 & \makecell[l]{0\\ $0< \cdot \leq 2000$\\$ 2000 < \cdot$}&\makecell[c]{0\\1\\2} & \makecell[c]{6,888\\2,281\\2,259}\\
    \midrule
    \ljpcriminal-Lv1 &\makecell{3\\3} &\makecell[l]{imprisonment with labor,\\imprisonment without labor} & month& \makecell[l]{0 \\ $0 < \cdot  \leq 6$ \\ $ 6 < \cdot $ } &\makecell[c]{0\\1\\2} & \makecell[c]{16,121\\3,219\\3,516}   \\
    \midrule 
    \midrule 
    \ljpcriminal-Lv2 & 5 &fine & KRW 1000 & \makecell[l]{0\\ $0< \cdot < 1000$\\$ 1000\leq \cdot <3000$ \\ $3000 \leq \cdot <10000$ \\ $10,000 \leq \cdot$}&\makecell[c]{0\\1\\2\\3\\4} & \makecell[c]{6,888\\982\\1,344\\2,046\\168}\\
    \midrule
    \ljpcriminal-Lv2& \makecell{6\\6} &\makecell[l]{imprisonment with labor,\\imprisonment without labor} & month& \makecell[l]{0 \\ $0 < \cdot  < 6$ \\ $ 6 \leq \cdot < 12$\\$12 \leq \cdot < 18 $ \\ $18 \leq \cdot < 60 $\\$ 60 \leq \cdot $ } &\makecell[c]{0\\1\\2\\3\\4\\5} & \makecell[c]{16,121\\942\\3,769\\1,615\\408\\1}   \\
    \midrule
    \midrule
    \ljpcriminal-Lv3 & $\infty$ &fine & KRW 1000 & - & - \\
    \midrule
    \ljpcriminal-Lv3 & $\infty$ &\makecell[l]{imprisonment with labor,\\imprisonment without labor} & month & - & -    \\
    \midrule
    \midrule

    \ljpcivil-Lv1 & 3 & \makecell{the degrees of claim acceptance\\(=approved money / claimed money)}& - & \makecell[l]{0\\$0 < \cdot <1$\\1\\}& \makecell[c]{0\\1\\2} & \makecell[c]{2,862\\2,132\\87} 
    \\
    \midrule
    \midrule
    \ljpcivil-Lv2 & 13 & \makecell{the degrees of claim acceptance\\(=approved money / claimed money)}& - & \makecell[l]{
    $0 \leq \cdot <0.05$
    \\$0.05 \leq  \cdot <0.15$
    \\$0.15 \leq \cdot <0.25$
    \\$0.25 \leq \cdot <0.35$
    \\$0.35 \leq \cdot <0.45$
    \\$0.45 \leq \cdot <0.55$
    \\$0.55 \leq \cdot <0.65$
    \\$0.65 \leq \cdot <0.75$
    \\$0.75 \leq \cdot <0.85$
    \\$0.85 \leq \cdot <0.95$
    \\$0.95 \leq \cdot < 1.05$ $^\dagger$
    \\$1.05 \leq \cdot < 1.15$ $^\dagger$
    \\$1.15 \leq \cdot <1.25$ $^\dagger$
    }
    & \makecell[c]{
    0\\0.1\\0.2\\0.3\\0.4\\0.5\\0.6\\0.7\\0.8\\0.9\\1\\1.1\\1.2
    } & \makecell[c]{
     2,935\\183\\227\\250\\245\\294\\213\\229\\194\\160\\143\\7\\1
    } 
    \\

    \bottomrule
  \end{tabular}\par 
  }
  \begin{tablenotes}[para]
  \item $\dagger$: The ratio of approved money / claimed money can be larger than one when the interest of money is involved. 
  \end{tablenotes}
\end{table}
\endgroup

\subsection{Pretraining}

\begingroup
\setlength{\tabcolsep}{2pt} 
\renewcommand{\arraystretch}{0.8} 

\begin{table}[h!]
\scriptsize
  \caption{Statistics of the pretraining corpora}
  \label{tbl_pretraining_staticstics}
  \centering
  \begin{tabular}{llcc}
    \toprule
    
    Domain & Pretraining Corpus & \# Tokens & Size (GB) \\
    
    \midrule
    
    Wiki & Wikipedia & 0.166B & 1.49 \\
    New, Book & Modu & 4.330B & 37.18 \\
    Legal & LBox-Open & 0.263B & 2.1 \\
    
    \bottomrule
  \end{tabular}
\end{table}
\endgroup

\begingroup
\setlength{\tabcolsep}{2pt} 
\renewcommand{\arraystretch}{0.8} 

\begin{table}[h!]
\scriptsize
  \caption{The model configuration and the pre-training parameters of \oursm}
  \label{tbl_pretraining_configuration}
  \centering
  \begin{tabular}{lccccccc}
    \toprule
    Name & $n_{\text{params}}$ & $n_{\text{layers}}$ & $d_{\text{model}}$ & $n_\text{head}$ & $d_\text{head}$ & Batch Size & Learning Rate  \\
    \midrule
    \oursm-base & 124M & 12 & 768 & 12 & 64 & 512 & $6.0 \times 10^{-4}$ \\
    \oursm-medium & 354M & 24 & 1024 & 16 & 64 & 512 & $6.0 \times 10^{-4}$  \\
    \bottomrule
  \end{tabular}
\end{table}
\endgroup

\begingroup
\setlength{\tabcolsep}{2pt} 
\renewcommand{\arraystretch}{0.8} 

\begin{table}[h!]
\centering

\tiny
  \caption{Comparison of various models on \ljpcriminal\ task under regression setting.
  }
  \label{tbl_baseline_ljp_criminal_regression}
  \centering
  \begin{threeparttable}
  \begin{tabular}{ll|lll}
    \toprule
    \multicolumn{1}{c}{Name} &
    \multicolumn{1}{c}{\makecell{size}} &
    \multicolumn{3}{c}{\makecell{regression\\ ($F_1$ )}} 
    \\
      &  
      & fine & \makecell{imp. \\w/ labor} & \makecell{imp.  \\w/o labor}  \\
    \midrule
    KoGPT-2-base   & 125M

    & $19.2$ & $51.6$ & $36.6$  
    \\    
    \oursm-base-zero & 124M

    & $21.6$ & $51.5$ & $36.0$  \\    
    mt5-small & 300M

    &$22.3$ & $51.8$ & $39.7$
    \\
    mt5-large (512 g)& 1.2B

    & $22.1$ &$52.1$  &$36.5$
    \\
    \midrule 
    KoGPT-2-base + d.a.  & 125M

    & $23.7$ & $51.5$ & $36.6$ 
    \\
    \oursm-base   & 124M

    & $20.0$ & $49.9$ & $39.1$ \\
    \oursm-base + d.a. & 124M

    & $25.8$& $52.0$& $40.0$ 
    \\
    \midrule
    \oursm-medium  & 354M

    & $21.8$ & $52.1$ &$36.0$ \\
    \oursm-medium + d.a. & 354M

    &$24.7$ & $53.5$ &$34.3$
    
    \\

    mt5-small + d.a.  & 300M

    & $25.4$ &$53.3$ &$36.6$ 
    \\
    \midrule
    Most frequent label & -

    &$12.7$ &$27.9$ &$5.4$  
    \\
    \midrule 
    Null ratio& -

    & 0.605 & 0.486 & 0.916 
    \\  
    Top non-null label ratio & -

    & 0.068 & 0.162 & 0.028 
    \\
    Top non-null label ratio (w/o null) & -

    &0.172 &0.314 &0.333 
    \\

    \bottomrule
  \end{tabular}
  \begin{tablenotes}[para]
  $\dagger$ The reason for the sentencing. 
  \end{tablenotes}
  \end{threeparttable}
\end{table}
\endgroup

\subsection{Author’s Statement}
We guarantee that \ours\ and \oursm\ are released under Attribution-NonCommercial-NoDerivatives 4.0 lincese and publicly available for non-commercial use.
We also use only the officially anonymyzed precedents issued by the Korean governments during the construction of the datasets to avoid possible confidentiality issues.
\ours, \oursm,  and the code to reproduce baseline results are available from \url{https://github.com/lbox-kr/lbox-open}.

\begingroup

\newpage
\begin{multicols}{2}

\subsection{Datasheet for the dataset}

\newcommand{\dssectionheader}[1]{%
   \noindent\framebox[\columnwidth]{%
      {\fontfamily{phv}\selectfont \textbf{{#1}}}
   }
}

\newcommand{\dsquestion}[1]{%
    {\noindent \fontfamily{phv}\selectfont {\textbf{#1}}}
}

\newcommand{\dsquestionex}[2]{%
    {\noindent \fontfamily{phv}\selectfont {\textbf{#1} #2}}
}

\newcommand{\dsanswer}[1]{%
   {\noindent #1 \medskip}
}

\dssectionheader{Motivation}

\dsquestionex{For what purpose was the dataset created?}{Was there a specific task in mind? Was there a specific gap that needed to be filled? Please provide a description.}

\dsanswer{
The datasets were created to stimulate research on natural legal language understanding tasks and to develop the real world machine learning applications in legal domain. The number of available datasets in legal domain is small especially for non-English language. Thus, we release the first large-scale benchmark of Korean legal AI datasets. 
}

\dsquestion{Who created this dataset (e.g., which team, research group) and on behalf of which entity (e.g., company, institution, organization)?}

\dsanswer{
    The dataset is created by the authors of this papers from LBox Co. Ltd. and KAIST.
}

\dsquestionex{Who funded the creation of the dataset?}{If there is an associated grant, please provide the name of the grantor and the grant name and number.}

\dsanswer{
    LBox Co. Ltd.
}

\dsquestion{Any other comments?}
\dsanswer{
None.}

\bigskip
\dssectionheader{Composition}

\dsquestionex{What do the instances that comprise the dataset represent (e.g., documents, photos, people, countries)?}{ Are there multiple types of instances (e.g., movies, users, and ratings; people and interactions between them; nodes and edges)? Please provide a description.}

\dsanswer{
The instances represent individual Korean precedents structured via our custom data engineering pipeline (Fig. \ref{fig_pipeline}.
}

\dsquestion{How many instances are there in total (of each type, if appropriate)?}

\dsanswer{
    The dataset statistics are shown in Table \ref{tbl_precedent_corpus} and \ref{tbl_tasks}.
}

\dsquestionex{Does the dataset contain all possible instances or is it a sample (not necessarily random) of instances from a larger set?}{ If the dataset is a sample, then what is the larger set? Is the sample representative of the larger set (e.g., geographic coverage)? If so, please describe how this representativeness was validated/verified. If it is not representative of the larger set, please describe why not (e.g., to cover a more diverse range of instances, because instances were withheld or unavailable).}

\dsanswer{
    The datasets are the sub set of entire Korean precedents whose their numbers are about few thousands million\footnote{\url{https://www.scourt.go.kr/portal/justicesta/JusticestaListAction.work?gubun=10}}. 
    As social phenomena and related legal aspects are so diverge, it is difficult to construct the representative sub set. 
    Thus, in \casename, \statute, \ljpcriminal, and \ljpcivil\ datasets, we used only the precedents (1) from 1st trials and (2) from limited case categories to construct the datasets.
    Under these constraints, we randomly sampled the precedents and the resulting datasets may approximately represent the parent distribution.
}

\dsquestionex{What data does each instance consist of? “Raw” data (e.g., unprocessed text or images) or features?}{In either case, please provide a description.}

\dsanswer{
    The instances of \ours\ consist of texts extracted from the Korean precedents except the labels of \ljpcriminal\  and \ljpcivil.
    The labels are generated by parsing and quantizing the numbers from the raw text that include either amount of claimed money, fine amount, or imprisonment period.
}

\dsquestionex{Is there a label or target associated with each instance?}{If so, please provide a description.}

\dsanswer{
Yes. See previous question. Also, see Table \ref{tbl_tasks} and Appendix \ref{sec: lbox-open-examples}.
}

\dsquestionex{Is any information missing from individual instances?}{If so, please provide a description, explaining why this information is missing (e.g., because it was unavailable). This does not include intentionally removed information, but might include, e.g., redacted text.}

\dsanswer{
    No.
}

\dsquestionex{Are relationships between individual instances made explicit (e.g., users’ movie ratings, social network links)?}{If so, please describe how these relationships are made explicit.}

\dsanswer{
    No.
}

\dsquestionex{Are there recommended data splits (e.g., training, development/validation, testing)?}{If so, please provide a description of these splits, explaining the rationale behind them.}

\dsanswer{
    Yes. See Table \ref{tbl_tasks}.
}

\dsquestionex{Are there any errors, sources of noise, or redundancies in the dataset?}{If so, please provide a description.}

\dsanswer{
    To make \ljpcriminal, we parsed the ruling to extract the fine amount and the imprisonment period using the custom language model. Although we filtered out the results with the confidence score from the language model and with a set of heuristically rules, the resulting parses could include imprecise numbers. Similarly, in \ljpcivil, the parsed monies from the gist of claim and the ruling could include noises. For the quality assurance, we manually checked around 40 examples from each dataset but found no errors. 
}

\dsquestionex{Is the dataset self-contained, or does it link to or otherwise rely on external resources (e.g., websites, tweets, other datasets)?}{If it links to or relies on external resources, a) are there guarantees that they will exist, and remain constant, over time; b) are there official archival versions of the complete dataset (i.e., including the external resources as they existed at the time the dataset was created); c) are there any restrictions (e.g., licenses, fees) associated with any of the external resources that might apply to a future user? Please provide descriptions of all external resources and any restrictions associated with them, as well as links or other access points, as appropriate.}

\dsanswer{
The datasets are self-contained.
}

\dsquestionex{Does the dataset contain data that might be considered confidential (e.g., data that is protected by legal privilege or by doctor-patient confidentiality, data that includes the content of individuals non-public communications)?}{If so, please provide a description.}

\dsanswer{
    No. Confidential information is anonymized by Korean government except some special cases where the governments decide to reveal for the social benefit. 
}

\dsquestionex{Does the dataset contain data that, if viewed directly, might be offensive, insulting, threatening, or might otherwise cause anxiety?}{If so, please describe why.}

\dsanswer{
The instances from the criminal cases may include the detailed description of crimes. Please see ``Ethical Consideration" section in the main paper.
}

\dsquestionex{Does the dataset relate to people?}{If not, you may skip the remaining questions in this section.}

\dsanswer{
    Yes.
}

\dsquestionex{Does the dataset identify any subpopulations (e.g., by age, gender)?}{If so, please describe how these subpopulations are identified and provide a description of their respective distributions within the dataset.}

\dsanswer{
The genders and ages may be implicitly identifiable from the facts that are included in the instances. But as they are written in free-form text, they are not easily identified without developing the dedicated parser. 
}

\dsquestionex{Is it possible to identify individuals (i.e., one or more natural persons), either directly or indirectly (i.e., in combination with other data) from the dataset?}{If so, please describe how.}

\dsanswer{
    No. Such information is anonymized by Korean government except some special cases where the government decide to reveal for the social benefit. 
}

\dsquestionex{Does the dataset contain data that might be considered sensitive in any way (e.g., data that reveals racial or ethnic origins, sexual orientations, religious beliefs, political opinions or union memberships, or locations; financial or health data; biometric or genetic data; forms of government identification, such as social security numbers; criminal history)?}{If so, please provide a description.}

\dsanswer{
    The instances from the criminal cases may include the detailed description of crimes, which can include sensitive data, and criminal history but the individuals are anonymized. Please see ``Ethical Consideration" section in the main paper.
}

\dsquestion{Any other comments?}
\dsanswer{
None.
}

\bigskip
\dssectionheader{Collection Process}

\dsquestionex{How was the data associated with each instance acquired?}{Was the data directly observable (e.g., raw text, movie ratings), reported by subjects (e.g., survey responses), or indirectly inferred/derived from other data (e.g., part-of-speech tags, model-based guesses for age or language)? If data was reported by subjects or indirectly inferred/derived from other data, was the data validated/verified? If so, please describe how.}

\dsanswer{
    All data are directly observable except the claim acceptance level in \ljpcivil\ and the fine and imprisonment levels in \ljpcriminal. For these cases, we include corresponding raw data, the gist of claim and the ruling.
}

\dsquestionex{What mechanisms or procedures were used to collect the data (e.g., hardware apparatus or sensor, manual human curation, software program, software API)?}{How were these mechanisms or procedures validated?}

\dsanswer{
The detailed procedure about how the dataset was constructed is described in Section \ref{sec: structuring} and Fig. \ref{fig_pipeline}.
}

\dsquestion{If the dataset is a sample from a larger set, what was the sampling strategy (e.g., deterministic, probabilistic with specific sampling probabilities)?}

\dsanswer{
    We selected precedents from  most frequent case categories. The detailed information is described in Section \ref{sec: datasets}. 
}

\dsquestion{Who was involved in the data collection process (e.g., students, crowdworkers, contractors) and how were they compensated (e.g., how much were crowdworkers paid)?}

\dsanswer{
The employees of LBox worked together. 
}

\dsquestionex{Over what timeframe was the data collected? Does this timeframe match the creation timeframe of the data associated with the instances (e.g., recent crawl of old news articles)?}{If not, please describe the timeframe in which the data associated with the instances was created.}

\dsanswer{
The dataset was created by buildling dedicated data engineering pipeline (Fig. \ref{fig_pipeline}) and took around an year. Without considering the pipeline building time, the process took around six monthes.
}

\dsquestionex{Were any ethical review processes conducted (e.g., by an institutional review board)?}{If so, please provide a description of these review processes, including the outcomes, as well as a link or other access point to any supporting documentation.}

\dsanswer{
The precedents are issued from the Korean courts and used to fill the empty space of the laws. Also, as they are mostly anonymized except some special cases, we did not take additional ethical review processes.
}

\dsquestionex{Does the dataset relate to people?}{If not, you may skip the remaining questions in this section.}

\dsanswer{
Yes.
}

\dsquestion{Did you collect the data from the individuals in question directly, or obtain it via third parties or other sources (e.g., websites)?}

\dsanswer{
We used the precedents issued by the Korean governments.
}

\dsquestionex{Were the individuals in question notified about the data collection?}{If so, please describe (or show with screenshots or other information) how notice was provided, and provide a link or other access point to, or otherwise reproduce, the exact language of the notification itself.}

\dsanswer{
No (See previous question).
}

\dsquestionex{Did the individuals in question consent to the collection and use of their data?}{If so, please describe (or show with screenshots or other information) how consent was requested and provided, and provide a link or other access point to, or otherwise reproduce, the exact language to which the individuals consented.}

\dsanswer{
No (see previous question).
}

\dsquestionex{If consent was obtained, were the consenting individuals provided with a mechanism to revoke their consent in the future or for certain uses?}{If so, please provide a description, as well as a link or other access point to the mechanism (if appropriate).}

\dsanswer{
N/A.
}

\dsquestionex{Has an analysis of the potential impact of the dataset and its use on data subjects (e.g., a data protection impact analysis) been conducted?}{If so, please provide a description of this analysis, including the outcomes, as well as a link or other access point to any supporting documentation.}

\dsanswer{
N/A.
}

\dsquestion{Any other comments?}

\dsanswer{
None.
}

\bigskip
\dssectionheader{Preprocessing/cleaning/labeling}

\dsquestionex{Was any preprocessing/cleaning/labeling of the data done (e.g., discretization or bucketing, tokenization, part-of-speech tagging, SIFT feature extraction, removal of instances, processing of missing values)?}{If so, please provide a description. If not, you may skip the remainder of the questions in this section.}

\dsanswer{
    Yes. Please refer to Section \ref{sec: datasets}.
}

\dsquestionex{Was the “raw” data saved in addition to the preprocessed/cleaned/labeled data (e.g., to support unanticipated future uses)?}{If so, please provide a link or other access point to the “raw” data.}

\dsanswer{
   When the raw data is required to reconstruct the processed data, we include them (for instance, in \ljpcivil\ and \ljpcriminal, we include the raw text from which the labels were generated).
}

\dsquestionex{Is the software used to preprocess/clean/label the instances available?}{If so, please provide a link or other access point.}

\dsanswer{
No.
}

\dsquestion{Any other comments?}

\dsanswer{
    None.
}

\bigskip
\dssectionheader{Uses}

\dsquestionex{Has the dataset been used for any tasks already?}{If so, please provide a description.}

\dsanswer{
    No. 
}

\dsquestionex{Is there a repository that links to any or all papers or systems that use the dataset?}{If so, please provide a link or other access point.}

\dsanswer{
    No.
}

\dsquestion{What (other) tasks could the dataset be used for?}

\dsanswer{
    The datasets can be used for (1) to train legal language models, (2) to train and benchmark machine learning models on natural legal language understanding tasks.
}

\dsquestionex{Is there anything about the composition of the dataset or the way it was collected and preprocessed/cleaned/labeled that might impact future uses?}{For example, is there anything that a future user might need to know to avoid uses that could result in unfair treatment of individuals or groups (e.g., stereotyping, quality of service issues) or other undesirable harms (e.g., financial harms, legal risks) If so, please provide a description. Is there anything a future user could do to mitigate these undesirable harms?}

\dsanswer{
The facts in the datasets can include personal information such as genders, ages, and regions. This information can bias the machine learning model trained using the datasets \citet{chalkidis2022facl_fairlex}. The users should be aware of this general properties of any data-driven approach and be cautious in interpreting the prediction of the model especially in the legal judgement prediction tasks.
}

\dsquestionex{Are there tasks for which the dataset should not be used?}{If so, please provide a description.}

\dsanswer{
    The datasets only partially cover various social phenomena and their related legal aspects. Because of this limitation and the reason explained in previous question, the trained models can generate imprecise and biased outputs. Thus the datasets should be used for academic purpose only.
}

\dsquestion{Any other comments?}
\dsanswer{None.}
\bigskip
\dssectionheader{Distribution}

\dsquestionex{Will the dataset be distributed to third parties outside of the entity (e.g., company, institution, organization) on behalf of which the dataset was created?}{If so, please provide a description.}

\dsanswer{
Yes, the datasets are publicly available.
}

\dsquestionex{How will the dataset will be distributed (e.g., tarball on website, API, GitHub)}{Does the dataset have a digital object identifier (DOI)?}

\dsanswer{
    The datasets \ours\ and pre-trained legal language model \oursm\ can be downloaded from HuggingFace hub (\ours: \url{https://huggingface.co/datasets/lbox/lbox_open}, \oursm: \url{https://huggingface.co/lbox/lcube-base/tree/main}).
}

\dsquestion{When will the dataset be distributed?}

\dsanswer{
    The datasets are currently available.
}

\dsquestionex{Will the dataset be distributed under a copyright or other intellectual property (IP) license, and/or under applicable terms of use (ToU)?}{If so, please describe this license and/or ToU, and provide a link or other access point to, or otherwise reproduce, any relevant licensing terms or ToU, as well as any fees associated with these restrictions.}

\dsanswer{
    The dataset are distributed under \texttt{Attribution-NonCommercial-NoDerivatives 4.0} license.
}

\dsquestionex{Have any third parties imposed IP-based or other restrictions on the data associated with the instances?}{If so, please describe these restrictions, and provide a link or other access point to, or otherwise reproduce, any relevant licensing terms, as well as any fees associated with these restrictions.}

\dsanswer{
    No. 
}

\dsquestionex{Do any export controls or other regulatory restrictions apply to the dataset or to individual instances?}{If so, please describe these restrictions, and provide a link or other access point to, or otherwise reproduce, any supporting documentation.}

\dsanswer{
    No.
}

\dsquestion{Any other comments?}

\dsanswer{
    None.
}

\bigskip
\dssectionheader{Maintenance}

\dsquestion{Who will be supporting/hosting/maintaining the dataset?}

\dsanswer{
    Wonseok Hwang, Dongjun Lee, and Kyoungyeon Cho will be maintaining/supporting the datasets.
}

\dsquestion{How can the owner/curator/manager of the dataset be contacted (e.g., email address)?}

\dsanswer{
    Via emails.
}

\dsquestionex{Is there an erratum?}{If so, please provide a link or other access point.}

\dsanswer{
    None found yet.
}

\dsquestionex{Will the dataset be updated (e.g., to correct labeling errors, add new instances, delete instances)?}{If so, please describe how often, by whom, and how updates will be communicated to users (e.g., mailing list, GitHub)?}

\dsanswer{
    Yes. The possible labeling errors, additional examples, and new tasks will be updated and announced via the LBox Open GitHub repository. The update interval is expected to be few months.
}

\dsquestionex{If the dataset relates to people, are there applicable limits on the retention of the data associated with the instances (e.g., were individuals in question told that their data would be retained for a fixed period of time and then deleted)?}{If so, please describe these limits and explain how they will be enforced.}

\dsanswer{
    No. 
}

\dsquestionex{Will older versions of the dataset continue to be supported/hosted/maintained?}{If so, please describe how. If not, please describe how its obsolescence will be communicated to users.}

\dsanswer{
    Yes. To maintain older version, we use HuggingFace hub (\url{https://huggingface.co}).
}

\dsquestionex{If others want to extend/augment/build on/contribute to the dataset, is there a mechanism for them to do so?}{If so, please provide a description. Will these contributions be validated/verified? If so, please describe how. If not, why not? Is there a process for communicating/distributing these contributions to other users? If so, please provide a description.}

\dsanswer{
    Yes. They can open the issue in the LBox Open GitHub repository.
}

\dsquestion{Any other comments?}

\dsanswer{
    None.
}
\end{multicols}

\end{document}